\documentclass[sigconf]{aamas}

\usepackage{balance} 
\usepackage{latexsym}
\usepackage{amsmath}
\usepackage{amsthm}
\usepackage{booktabs}
\usepackage{enumitem}
\usepackage{graphicx}
\usepackage{color}

\usepackage{subcaption}
\usepackage{multirow}
\usepackage{tabu}
\usepackage{placeins}

\usepackage{algorithm}
\usepackage{algorithmic}

\newcommand{\PP}[2][]{\mathbb{P}_{#1}\left[#2\right]}   
\newcommand{\EE}[1]{\mathbb{E}_{\pi_\theta}\left[#1\right]}   
\newcommand{\ET}[1]{\mathbb{E}_{\tau\sim\mathbb{P}(\cdot|\theta)}\left[ #1\right]}  
\newcommand{\ETi}[1]{\mathbb{E}_{\tau^i\sim\mathbb{P}(\cdot|\theta)}\left[ #1\right]}  

\newcommand{\TDR}[1]{\mathbb{E}_{\substack{s_0\sim\mu(\cdot)\\a\sim\pi_\theta(\cdot|s)}}\left[#1\right]}
\newcommand{\loss}[1]{\mathcal{L}\big(#1\big)} 
\newcommand{\RR}{\mathbb{R}} 

\newcommand{\ie}{\emph{i.e.}}
\newcommand{\eg}{\emph{e.g.}}

\newcommand{\Ss}{\mathcal{S}}
\newcommand{\Aa}{\mathcal{A}}

\newcommand{\Ee}{\mathcal{E}}
\newcommand{\Gg}{\mathcal{G}}
\newcommand{\Nn}{\mathcal{N}}
\newcommand{\Mm}{\mathcal{M}}
\newcommand{\Pp}{\mathcal{P}}
\newcommand{\Oo}{\mathcal{O}}
\newcommand{\Tt}{\tau}
\newcommand{\CI}[2]{$\begin{array}{c}{#1}\\({#2})\end{array}$}  
\newcommand{\CB}[2]{$\begin{array}{c}\mathbf{{#1}}\\({#2})\end{array}$} 

\newcommand{\tabbreak}[2]{$\begin{array}{c}{\mathbf{#1}}\\\mathbf{#2}\end{array}$}

\setcopyright{ifaamas}
\acmConference[AAMAS '25]{Proc.\@ of the 24th International Conference
on Autonomous Agents and Multiagent Systems (AAMAS 2025)}{May 19 -- 23, 2025}
{Detroit, Michigan, USA}{Y.~Vorobeychik, S.~Das, A.~Nowe (eds.)}
\copyrightyear{2025}
\acmYear{2025}
\acmDOI{}
\acmPrice{}
\acmISBN{}

\acmSubmissionID{442}

\title[AAMAS-2025 Formatting Instructions]{Networked Agents in the Dark: Team Value Learning under Partial Observability}

\author{Guilherme S. Varela}
\affiliation{
  \institution{INESC-ID/Instituto Superior Técnico}
  \city{Lisbon}
  \country{Portugal}}
\email{guilherme.varela@tecnico.ulisboa.pt}

\author{Alberto Sardinha}
\affiliation{
  \institution{PUC-Rio}
  \city{Rio de Janeiro}
  \country{Brazil}}
\email{sardinha@inf.puc-rio.br}

\author{Francisco S. Melo}
\affiliation{
  \institution{INESC-ID/Instituto Superior Técnico}
  \city{Lisbon}
  \country{Portugal}}
\email{fmelo@inesc-id.pt}

\begin{abstract}
We propose a novel cooperative multi-agent reinforcement learning (MARL) approach for networked agents. In contrast to previous methods that rely on complete state information or joint observations, our agents must learn how to reach shared objectives under partial observability. During training, they collect individual rewards and approximate a team value function through local communication, resulting in cooperative behavior. To describe our problem, we introduce the networked dynamic partially observable Markov game framework, where agents communicate over a switching topology communication network. Our distributed method, DNA-MARL, uses a consensus mechanism for local communication and gradient descent for local computation. DNA-MARL increases the range of the possible applications of networked agents, being well-suited for real world domains that impose privacy and where the messages may not reach their recipients. We evaluate DNA-MARL across benchmark MARL scenarios. Our results highlight the superior performance of DNA-MARL over previous methods.
\end{abstract}


\newcommand{\BibTeX}{\rm B\kern-.05em{\sc i\kern-.025em b}\kern-.08em\TeX}

\begin{document}


\pagestyle{fancy}
\fancyhead{}


\maketitle 


\section{Introduction}

Cooperative multi-agent reinforcement learning involves rational agents learning how to behave under uncertainty in a shared environment to maximize a single utility function. While distributed training has once been the dominating paradigm in learning for MARL~\cite{busoniu_2010} systems, the community's focus has shifted to centralized training and decentralized execution (CTDE) in recent years. The fundamental reason is that centralized training agents benefit from a either a single loss function to train a common policy, e.g., parameter sharing~\cite{gupta_2017}, or from agent-wise policies that can be factorized between agents, e.g., Q-MIX~\cite{rashid_2018}.

Centralized training is ideal in settings where data is centralized. During training, agents benefit from sharing information, such as {\em joint observations} for partial observability mitigation, {\em joint actions} for modeling teammates and {\em team rewards} for cooperation. However, CTDE has its limitations. It assumes the existence of a central node (or entity) that actually trains the agents, knows all system information, performs all the necessary computations, and then distributes the resulting individual policies to agents for execution.

Distributed training with decentralized execution has re-emerged, e.g,~\cite{zhang_2018} and~\cite{chen_2022}, as an alternative to CTDE systems in real world domains where there is no central entity capable of performing computations in behalf of the agents. For instance, in scenarios like distributed economic dispatch~\cite{tu_2024}, where agents collaborate to determine optimal power generation, it is crucial to preserve the privacy of agents' observations--their power generation and cost curves. This privacy protection is essential for the fair bidding in the sale or purchase of energy. Hence agents collaborate to achieve a common goal, but they are less forthcoming about sharing their own observations. Another example of application is distributed packet routing in a dynamically changing networks~\cite{boyan_1993}. Agents are nodes and by using {\em local} observations and collecting {\em individual rewards}, must balance the selection of routes that minimize the number of "hops" of a given packet, against the risk of overflowing links along popular routes. Finally, in wireless sensor networks~\cite{kar_2010} agents are endowed with limited processing and communication capacities, precluding exchanges of high precision (analog) data. Furthermore, randomness in the environment results in random packet dropouts.

In this work, we advance upon the decentralized training and decentralized execution (DTDE)~\cite{gronauer_2022} paradigm, wherein {\em networked agents} use peer-to-peer communication during training and operate in isolation during execution. Prior work has produced networked agents under relaxed assumptions:~\citet{zhang_2018} assumed a fully observable state while the rewards are kept private, and~\citet{chen_2022} proposed networked agents that choose when and to whom request observations. In contrast, we introduce a novel approach that is not bound by the same restrictions and our agents learn under partial observability. The key to our method is the use of a consensus mechanism to force agents to agree on a team value, resulting in cooperative value function learning under the partial observability setting. 

In summary, our key contributions can be outlined as follows. First, we formalize the {\em networked dynamic partially observable Markov game} (ND-POMG), a specialized framework derived from {\em partially observable Markov game} where agents communicate over a switching topology network. Second, we present a novel approach, DNA-MARL, for solving ND-POMG problems with a team policy gradient. This approach is implemented in an actor-critic algorithm and extended to the deep $Q$-network algorithm, showing its generality. Finally, we evaluate our approach and show that it outperforms other decentralized training, decentralized execution systems.
\section{Background}

\textbf{Partially observable Markov game} (POMG): We define  a \textit{partially observable Markov game}~\cite{oliehoek_2016} for $N$ agents as the tuple:
\begin{equation*}
(\Nn, \Ss, \{\Aa^i\}_{i\in\Nn},\{\Oo^i\}_{i\in\Nn}, \Pp, \{r^i\}_{i\in\Nn}, \gamma)\text{,}
\end{equation*}
where $\Nn = \{1,\dots, N\}$ denotes a set of $N$ agents. $\Ss$ represents the {\em state space} describing the system, which is not observed. Instead, at each time step $t$, each agent $i\in \Nn$ observes $o^i_t\in \Oo^i$, that depends on the true system state $s\in \Ss$. The set of {\em actions} available to agent $i$ is denoted by $\Aa^i$. The joint action set $\Aa$ is the Cartesian product of the individual action spaces, \ie, $\Aa= \Aa^1\times\dots\times\Aa^N$. The transition probability $\Pp:\Ss\times\Aa\to\Delta(\Ss)$ denotes the probability distribution over the next state, they depend on the joint action $a\in\Aa$ and the current state $s$. The instantaneous individual reward for agent $i$ is given by $r^i:\Ss\times\Aa\to\mathbb{R}$; $\gamma \in [0, 1)$ is a discount factor.

\textbf{Actor-critic}: is a class of model-free reinforcement learning (RL)  algorithms aimed at optimizing  the policy $\pi_\theta$, parameterized by $\theta\in\Theta$.  Particularly, the {\em actor} component updates the policy $\pi_\theta$,  and the {\em critic} component evaluates the actor's policy performance by using the $Q$-function:
\begin{equation*}
Q(s,a;\theta) =\EE{\sum^{\infty}_{k=t}\gamma^{k-t} r_{k+1}| s_t=s, a_t=a}\text{.}
\end{equation*}
The $Q$-function yields the expected discounted return by taking action $a$ on state $s$ at time $t$, and then following $\pi_\theta$ thereafter. In the single agent setting, the policy gradient theorem~\cite{sutton_1999} prescribes the direction for the gradient updates to maximize the {\em total discounted return}:  
\begin{equation}\label{eqn:PG}
    \nabla_\theta J(\pi_\theta)=\EE{\nabla_\theta \text{log } \big(\pi(a_t| s_t; \theta)\big) Q(s_t, a_t;\theta)}\text{,}
\end{equation}

\textbf{Actor-critic with advantage} (A2C)~\cite{dhariwal_2017}, in the single agent episodic setting,  the history of the interactions with the environment are collected into trajectories. A mini-batch is the concatenation of many trajectories, drawn from the same policy using parallel processing. A2C maintains one neural network for the actor, and one another for the critic, their weights are adjusted via gradient descent. The critic updates its parameters $\omega$ by minimizing the {\em least mean squares} loss function:
\begin{equation}\label{eqn:critic_update}
\loss{\omega; \Tt}= \frac{1}{T}\sum_{(s, a)\in \Tt}||A(s, a; \omega)||^2_2\text{,}
\end{equation}
where $T$ is the length of an episode,  $A(s, a;\omega) = Q(s, a;\omega) - V(s;\omega)$ is the {\em advantage function}, and the value function
\begin{equation*}
    V(s; \omega) =\EE{\sum^{T}_{k=t} \gamma^{k-t} r_{k+1}| s_t=s, \omega}\text{,}
\end{equation*}
captures the discounted return for being on state $s$ at time $t$, and then following $\pi_\theta$ thereafter until the episode's end at $T$. The advantage function reduces the variance of the actor-critic gradient updates. The actor updates its parameters $\theta$ by minimizing the loss function:
\begin{equation}\label{eqn:actor_update}
\loss{\theta; \Tt} = -\sum_{(s, a)\in \Tt}\text{log}\big(\pi(a|s;\theta)\big)A(s, a; \omega)\text{.}
\end{equation}

\textbf{Consensus}: The goal of randomized consensus algorithms is to asymptotically reach an agreement on the global average of individual parameters held by nodes in a switching topology communication network through local communication. Formally, the switching topology communication network is defined by an undirected graph $\Gg_k(\Nn,\Ee_k)$, where $\Nn=\allowdisplaybreaks \{1,\dots, N\}$ is the node set, and $\Ee_k\subseteq \Nn\times\Nn$ denotes the time-varying edge set with respect to communication step $k$\footnote{In this work $t$ represents timesteps in episodic interactions with the environment, while $k$ represents communication timesteps (rounds) that occur between episodes. Communication only happens during training, in between episodes. During execution agents are {\em fully decentralized}.}. Nodes $n$ and $m$ can communicate at communication step $k$, if and only if, $(n, m) \in \Ee_k$. Each node $n$, initially holding a parameter $\phi^n(0)$, has the opportunity at each communication step $k$, to synchronously interact with its neighbors, updating its parameter value by replacing its own parameter with the average of its parameter and the parameters from neighbors. The distributed averaging consensus algorithm~\cite{xiao_2007} prescribes the updates:
\begin{equation}\label{eqn:consensus}
  \phi^n(k+1) = \sum_{m\in\Nn^n_k} W^{n, m}_k \cdot \phi^m(k)\text{,} 
\end{equation}
where $\Nn^n_k=\{m|(n, m)\in\Ee_k\}$ represents the neighborhood of agent $n$ at time $k$. For a switching topology dynamic with random link dropouts, it is possible to show that in the limit, the values of the parameters for each node $n$ converge to the network's average, \ie:
 \begin{equation}\label{eqn:limit}
 \lim_{k\to \infty}\phi^n(k) =\frac{1}{N}\sum^N_n\phi^n(0)\text{.}
 \end{equation}
Moreover, for an arbitrary graph $\Gg_k(\Nn, \Ee_k)$, it is possible to derive the weights $W^{n,m}_k$ that guarantee consensus locally. For instance, the {\em Metropolis weights} matrix~\cite{xiao_2007} in (Appendix~\ref{appendix:metropolis}) is a matrix that guarantee consensus, requiring only that each node be aware of its closest neighbor degree.

 \textbf{Networked agents} is a class of distributed reinforcement learning agents that combines consensus iterations in~\eqref{eqn:consensus} for localized approximations and actor-critic updates in~\eqref{eqn:actor_update} and~\eqref{eqn:critic_update}. Relevant previous works include:
 
\textbf{Critic consensus}:~\citet{zhang_2018} introduce networked agents where the critic network $V(\cdot, \cdot;\omega)$, parameterized with $\omega$, approximates the value-function $V^\pi(\cdot)$. The distributed critic emulates a central critic. Agents observe the transition $(s_t, a_t, s_{t+1})$, perform critic the update in~\eqref{eqn:critic_update}, then agents average the parameter using consensus:
\begin{equation}\label{eqn:omega_consensus}
   \omega^i(k+1) = \sum_{j\in\Nn^i_k} W^{i,j}_k\cdot \omega^j(k)\quad \forall i\in \Nn\text{.}
\end{equation} 

\textbf{Policy consensus}:~\citet{chen_2022} introduce the class of homogeneous Markov games wherein there is no suboptimality incurred by performing consensus on the actor parameters. Their motivation is to emulate parameter sharing under the decentralized setting, while minimizing the number of communication rounds. Agents perform the actor update in~\eqref{eqn:actor_update}, then average the parameters using consensus:
\begin{equation}\label{eqn:theta_consensus}
   \theta^i(k+1) = \sum_{j\in\Nn^i_k} W^{i,j}_k\cdot \theta^j(k)\quad \forall i\in \Nn\text{.}
\end{equation} 

 \section{Networked Dynamic POMG}
 
 In this section, we present the first key contribution, which is a formalization of {\em networked dynamic partially observable Markov game}, ND-POMG. We define the ND-POMG as the septuple:
\begin{equation*} \Mm=(\Gg_k,\Ss, \{\Oo^i\}_{i\in\Nn}, \{\Aa^i\}_{i\in\Nn}, \Pp, \{r^i\}_{i\in\Nn}, \gamma)\text{,} \end{equation*}
where $\Gg_k(\Nn, \Ee_k)$ represents a switching topology communication network, and the latter six elements represent the POMG elements.

In this work, we fix the agents set $N=|\Nn|$, to ensure that no agent is added or removed from the network. We also introduce the hyperparameter $C=|\Ee_k|$ for all $k$, that shapes the topology of the communication network by fixing the cardinality of every possible edge set $\Ee_k$. Moreover, we let $\Ee_k$  change according to an uniform distribution at each communication round.  The uniform distribution over the edge sets is the least specific distribution that guarantees that over a sufficiently long round of communications agents will reach consensus (Appendix~\ref{appendix:graph_model}).

\section{Double Networked Averaging MARL}\label{sec:DNAMARL}

This section presents our second key contribution which is the DNA-MARL an approach to solve ND-POMG problems. Since our method requires an extra consensus iteration step, we call it {\em double networked averaging} MARL (DNA-MARL). Any single agent reinforcement learning algorithm can be cast as a DNA-MARL with our method, we elaborate the case for the A2C, an {\em on-policy}  method (Sec.~\ref{sec:DNAA2C}) and extend to the deep $Q$-network (DQN) (Sec.~\ref{sec:DNAQL}), an {\em off-policy} method.
 \subsection{Double Networked Averaging A2C}\label{sec:DNAA2C}
 
In order to make agents cooperate with decentralized training, we  factorize the shared objective between agents. Hence, it is possible to maximize performance via local communication and local gradient descent updates.

The {\em total discounted team return}, $J(\pi_\theta)$,  serves as a  measure of the joint policy $\pi_\theta$  performance:
 \begin{equation}
 J(\theta) = \TDR{\sum^{T}_{t=0}\gamma^t r_{t+1}} \text{,}      
 \end{equation}\label{eqn:TDR} 
where $r_{t+1}=\frac{1}{N}\sum_{i\in\Nn} r^i_{t+1}$ is the {\em instantaneous team reward}. The expectation is taken by drawing the initial state $s_0$ from the initial state distribution $\mu$ and taking actions from $\pi_\theta$,  thereafter. For simplicity, we follow the convention of writing $J(\pi_\theta)$ as $J(\theta)$.

\subsubsection{Team Policy Gradient}

To obtain the team policy gradient, we replace the team reward in~\eqref{eqn:TDR} by the average of {\em  individual rewards}.

\begin{align*}
      J(\theta) &=  \TDR{\sum^{T}_{t=0}\gamma^t \big(\frac{1}{N}\sum_{i\in\Nn}r^i_{t+1}\big)}\\
      & =\sum_{i\in\Nn}\TDR{ \sum^{T}_{t=0}\frac{\gamma^t}{N} r^i_{t+1}} =\sum_{i\in\Nn}J^i(\theta)
\end{align*}

 The result above suggests how the cooperative system's objective can be distributed across the participating agents, thus the total discounted team return is computed as the weighted sum of the discounted individual rewards. However,  the behaviors of the agents are still coupled, depending on the joint policy parameterized by $\theta\in\Theta$ and on the  common system state $s_t$. Formally, the objective of the cooperative distributed system is to maximize the {\em total discounted team return}:
\begin{equation}\label{eqn:objective}
    \max_\theta \sum_{i\in\Nn}J^i(\theta)\text{ with }J^i(\theta)=\mathbb{E}_{\Tt\sim\mathbb{P}(\cdot|\theta)}\left[ \sum^{T}_{t=0}\gamma^t r^i_{t+1}\right]\text{.}
\end{equation}
We drop the scaling constant $N$ as it does not change the stationary points of the maximization. $\mathbb{P}(\cdot|\theta)$ is the short hand notation for the probability distribution of the trajectories, 
\begin{equation*}
\Tt=\big(s_0, a_0, r_1, s_1, a_1, r_2, \dots, s_T\big)\text{,}
\end{equation*}
generated from system dynamics $\Pp$ under the joint policy $\pi_\theta(s)$.
The maximization can be achieved through iterative gradient search methods. More specifically, the policy gradient theorem \eqref{eqn:PG} prescribes the direction of the parameter updates:
\begin{equation*}
\nabla_\theta J(\theta) = \ET{\nabla_\theta \text{log }\big(\pi_\theta(a|s)\big)A_\theta(s, a)}\text{,}
\end{equation*}
that maximize the total discounted team return. We replaced the $Q$-function by the {\em advantage function} \eqref{eqn:critic_update} to mitigate the variance on the weight updates. We note that the single agent policy gradient update in \eqref{eqn:PG} serves as the policy gradient for a centralized agent in control of all agents. Departing from the centralized setting, we use the fact that the joint policy $\pi_\theta$ factorizes between agents, to set:
\begin{align}\label{eqn:grad_J}
\nabla_\theta J(\theta)  &=\ET{\nabla_\theta \text{log }\Big(\Pi_{i\in\Nn}\pi^i_{\theta^i}(a^i|s)\Big)A_\theta(s, a)}\nonumber\\
&= \ET{\Big(\sum_{i\in\Nn}\nabla_{\theta^i} \text{log }\big(\pi^i_{\theta^i}(a^i|s)\big)\Big)A_\theta(s, a)}\nonumber\\
&= \sum_{i\in\Nn}\ET{\nabla_{\theta^i} \text{log }\big(\pi^i_{\theta^i}(a^i|s)\big)A_\theta(s, a)}\text{.}
\end{align}
There are two limitations in \eqref{eqn:grad_J} preventing its use for conducting local updates. In the context of partial observability the states $s$ are unavailable in trajectory $\Tt$. Second, the gradient update depends on global $\theta\in\Theta$, and no agent has access to $\theta$ . 

\subsubsection{Distributed Reinforcement Learning}

We address the limitations in~\eqref{eqn:grad_J} by considering the {\em information structure} of the problem,  or what do agents know~\cite{zhang_2021}. We propose localized approximations that allow agents to perform local updates:  considering a synchronous system where agents interact with the environment to collect their individual trajectories $\Tt^i$. Distributed learning {\em requires} a localized approximation $\nabla_\theta J^i(\theta)$ for gradient of the discounted team return $\nabla_{\theta} J(\theta)$ in \eqref{eqn:grad_J}. Moreover, each agent maximizes its policy $\pi^i_\theta$ which is parameterized by $\theta^i$, \ie, $\pi^i_\theta = \pi^i_{\theta^i}$ and $\nabla_\theta J^i(\theta)=\nabla_{\theta^i} J^i(\theta)$. Combining the three facts together the localized approximation for \eqref{eqn:grad_J} can be rewritten as: 

\begin{equation}\label{eqn:grad_Ji}
\nabla_{\theta^i} J(\theta^i) = \ETi{\nabla_{\theta^i} \text{log }\Big(\pi^i_{\theta^i}(a^i|o^i)\Big)A_\theta^i(o^i, a^i)}\text{,}
\end{equation}
where $\Tt^i=\big(o^i_0, a^i_0, r^i_1, o^i_1, a^i_1, r^i_2, \dots, o^i_T\big)$ is available locally for agent $i$. The replacement $s^i_t$ with $o^i_t$ under the partially observability setting is standard practice in MARL literature~\cite{chen_2022, rashid_2018, sunehag_2018}. The system's dynamics still depend on the joint behavior, parameterized by $\theta$, but the gradient in \eqref{eqn:grad_Ji} is locally defined. Straightforward application of the actor-critic updates in \eqref{eqn:actor_update} and \eqref{eqn:critic_update}, with individual rewards over local parameters lead to {\em independent learners}, which evaluate their individual policies. Individual learners assume that the approximation $A^i_{\theta^i}(o^i, a^i) \approx A_\theta(o, a)$ holds.

\subsubsection{Distributed Cooperation}\label{sec:distributed_cooperation}

We propose a better approximation for the gradient of the discounted team return by performing the updates in the direction of the {\em team advantage} $A_\theta(s, a)$ in \eqref{eqn:grad_J}, rather than the local advantage $A_\theta^i(o^i, a^i)$ in \eqref{eqn:grad_Ji}. However, since the team advantage is unavailable, agents should instead perform local updates in the direction of the  {\em team advantage under the partially observable} setting $A_\theta(o, a)$. Since $o =\left[o^1,\dots, o^N\right]$ is the concatenation of observations, $A_\theta(o,a)$  can be defined by:
\begin{align}\label{eqn:team_A} A_\theta(o, a)  = Q(o, a;\omega) - V(o;\omega) \approx r + \gamma V(o';\omega_{-}) - V(o;\omega)\text{,}\end{align}
where, $r$  and  $o'$ are respectively the rewards and the joint observations on the next time step. The parameter $\omega_{-}$ is a periodic copy of the critic's parameters $\omega\in\Omega$, which serves to stabilize learning. Decentralized learning agents neither observe $o$ nor collect $r$,  but may resort to local communication schemes to obtain  factorized representations for $V(o, a;\omega)$. The local critic update is a straightforward adaptation of the single agent critic in~\eqref{eqn:critic_update}:
\begin{equation}\label{eqn:local_critic}
 \loss{\omega^i; \tau^i}= \frac{1}{T}\sum^{T-1}_{t=0}\big(y^i_t- V(o^i_t; \omega^i)\big)^2 \text{,}
\end{equation}
with
\begin{equation*}
y^i_t = r^i_{t+1} + \gamma V(o^i_{t+1}; \omega^i_{-})\text{.}
\end{equation*}
 We propose to  use communication to combine $y^i_t$ by performing team-$V$ consensus:
\begin{equation}\label{eqn:V_consensus}
    y^i_t(k+1) =  \sum_{j\in\Nn^i_k} W^{i, j}_k \cdot y^j_t(k) \quad \forall i\in \Nn, k=1,\dots, K\text{.} 
\end{equation}
After each training episode, agents concurrently approximate the team-$V$ using $y^i_t$ based on their individual rewards and observations. Then, we let $K$ consensus updates per mini-batch aimed at approximating the team-$V$. At each communication round, a connected agent averages its team-$V$ estimation with team-$V$s from neighbors. Ideally, the following approximation will hold:  
\begin{equation}\label{eqn:team_V}
    \bar{y}^i_t =  \sum_{i=1}\frac{1}{N}\left[r^i_{t+1} +  \gamma V(o^i_{t+1};\omega^i_{-})\right]  \approx \sum_{i=1} \frac{1}{N}V(o^i_t; \omega^i_{-})\text{.} 
\end{equation}
We empirically test for suitable values of $K$. The consensus steps in \eqref{eqn:V_consensus}  result in a flexible degree of cooperation: When $K=0$, agents behave as {\em  independent learning} agents.  For a high enough values of $K$, the approximation error should be small enough, such that agents behave in fully cooperative mode.

This section concludes a core contribution to our method: distributed cooperation  whereby agents produce localized approximations for a team-$V$ (or team-$Q$) using consensus.   Cooperation requires that each agent approximates the same critic, and this critic must evaluate the joint policy, so that the actor updates its parameters in the direction of the team-$V$,  thus the best local actions for the team will be reinforced. Moreover, the updates in~\eqref{eqn:team_V} do not require agents to be homogeneous. Previous networked agents works~\cite{zhang_2018, chen_2022} provide asymptotic convergence guarantees for linear function approximation on the state-action space. Under the linearity approximation on the critic and fully observable setting the update in~\eqref{eqn:omega_consensus} is sufficient to guarantee cooperation. Under partial observability and/or non-linear function observation agents are unable to obtain localized approximations for the team-$V$/team-$Q$ by the averaging of critic parameters .

\subsubsection{Algorithm}

To design our algorithm, we combine the localized approximations for the team-$V$ in~\eqref{eqn:team_V} with critic consensus in~\eqref{eqn:omega_consensus}~\cite{zhang_2018}. And  actor consensus in~\eqref{eqn:theta_consensus}~\cite{chen_2022} for improving sample efficiency. Hence, the updates comprising the double networked averaging actor critic with advantage are given  by:

\begin{align*}
 y^i_t(k+1) &=  \sum_{j\in\Nn^i_k} W^{i, j}_k \cdot y^j_t(k) \quad k=1,\dots, K\tag{i}\label{eqn:step_1_team_V}\\
\bar{y}^i_t &=  y^i_t(K+1)\tag{ii}\label{eqn:step_2_team_V}\\
\end{align*}
Every agent interacts locally with the environment to collect the individual trajectories $\Tt^i$. Then, in ~\eqref{eqn:step_1_team_V} agents use localized approximation for team-$V$, $\bar{y}^i_t$, by performing $K$ consensus steps; ~\eqref{eqn:step_2_team_V} the final approximation for the team-$V$ is defined; The next steps consist of local weight updates:

\begin{align*}
 \loss{\omega^i; \Tt^i, \bar{y}^i} &= \frac{1}{T}\sum^{T-1}_{t=0}\big(\bar{y}^i_t -  V(o^i_t; \omega^i)\big)^2\tag{iii}\label{eqn:step_3_team_critic}\\
\loss{\theta^i; \Tt^i, \bar{y}^i} &= -\frac{1}{T}\sum^{T-1}_{t=0}\text{log}\pi(a^i_t|o^i_t;\theta^i)( \bar{y}^i_t - V(o^i_t; \omega^i))\tag{iv}\label{eqn:step_4_team_actor}\\
\end{align*}
In~\eqref{eqn:step_3_team_critic}  the local critic updates its parameters using $\bar{y}^i_t$ instead of their own estimations~$y^i_t$; ~\eqref{eqn:step_4_team_actor} Similarly,  actor updates its parameters in the direction of  team-$V$; Finally, periodically agents perform actor and critic  parameter consensus, represented by  steps~\eqref{eqn:step_5_omega_consensus} and~\eqref{eqn:step_6_theta_consensus}:
\begin{align*}
    \omega^i(k+1) &=  \sum_{j\in\Nn^i_k} W^{i, j}_k \cdot \omega^j(k)\quad k=1,\dots, K\tag{v}\label{eqn:step_5_omega_consensus}\\
    \theta^i(k+1) &=  \sum_{j\in\Nn^i_k} W^{i, j}_k \cdot \theta^j(k)\quad k=1,\dots, K\tag{vi}\label{eqn:step_6_theta_consensus}
\end{align*}
We note that actor-critic parameters can be concatenated to avoid extra communication rounds, and that~\eqref{eqn:step_5_omega_consensus}  utilizes~\eqref{eqn:omega_consensus}  and that~\eqref{eqn:step_6_theta_consensus} utilizes \eqref{eqn:theta_consensus}.  And the Listing~\ref{alg:DNAA2C} in (Appendix~\ref{appendix:pseudocodes}) provides the pseudocode. The consensus updates in~\eqref{eqn:step_5_omega_consensus} require agents to have homogeneous observation spaces, while updates in~\eqref{eqn:step_6_theta_consensus} require agents to have homogeneous action spaces. 

Figure~\ref{fig:diagram} illustrates the information flow of the algorithm, clockwise from the left: (a) Four agents (circles) interact with the environment and evaluate a team-$V$ (blue gradient) from their individual experiences; (b) Consensus on team-$V$ occurs over the time varying communication network. At each step, certain agents aggregate their opinions on the team-$V$ \eqref{eqn:V_consensus}; (c) Agents independently update their parameters using gradient descent, resulting in varying actor-critic evaluations and policies (orange gradient); (d) Parameter consensus occurs periodically over the time varying communication network. At each step, certain agents aggregate their opinions on the parameters in (\cite{zhang_2018, chen_2022}). Pseudo codes for double networked agents are provided in (Appendix~\ref{appendix:pseudocodes}) and we open source the codebase \texttt{DNA-MARL}\footnote{\label{note:DNA-MARL}\texttt{\url{https://github.com/GAIPS/DNA-MARL}}}.

\subsection{Double Networked Averaging Q-learner}\label{sec:DNAQL}

 We extend the DNA method to the independent $Q$ learner. To compute the team-$Q$ consensus, which involves the averaging of individual $Q$-function evaluations, agents must first  determine locally the {\em learning target}, 
\begin{equation}\label{eqn:Q_local}y^i_t = r^i_{t+1} + \gamma\max_{a'\in\Aa^i}Q(o^i_{t+1}, a';\theta^i_{-})\text{.}\end{equation}

Similarly to single agent DQN~\cite{mnih_2015}, the learning target consists of the sum of the local reward $r^i_{t+1}$ and the $Q$-value assigned to the individual action $a'$ that yields the highest $Q$-value over the next observation $o^i_{t+1}$. Moving from the $V$-function in \eqref{eqn:local_critic}  to $Q$-function in \eqref{eqn:Q_local}, the learning target becomes a function of a $\max$ operator, which is performed locally. The consensus updates for the team-$Q$ are then calculated as follows: 

\begin{figure}[t]
    \centering
          \includegraphics[width=0.3\textwidth]{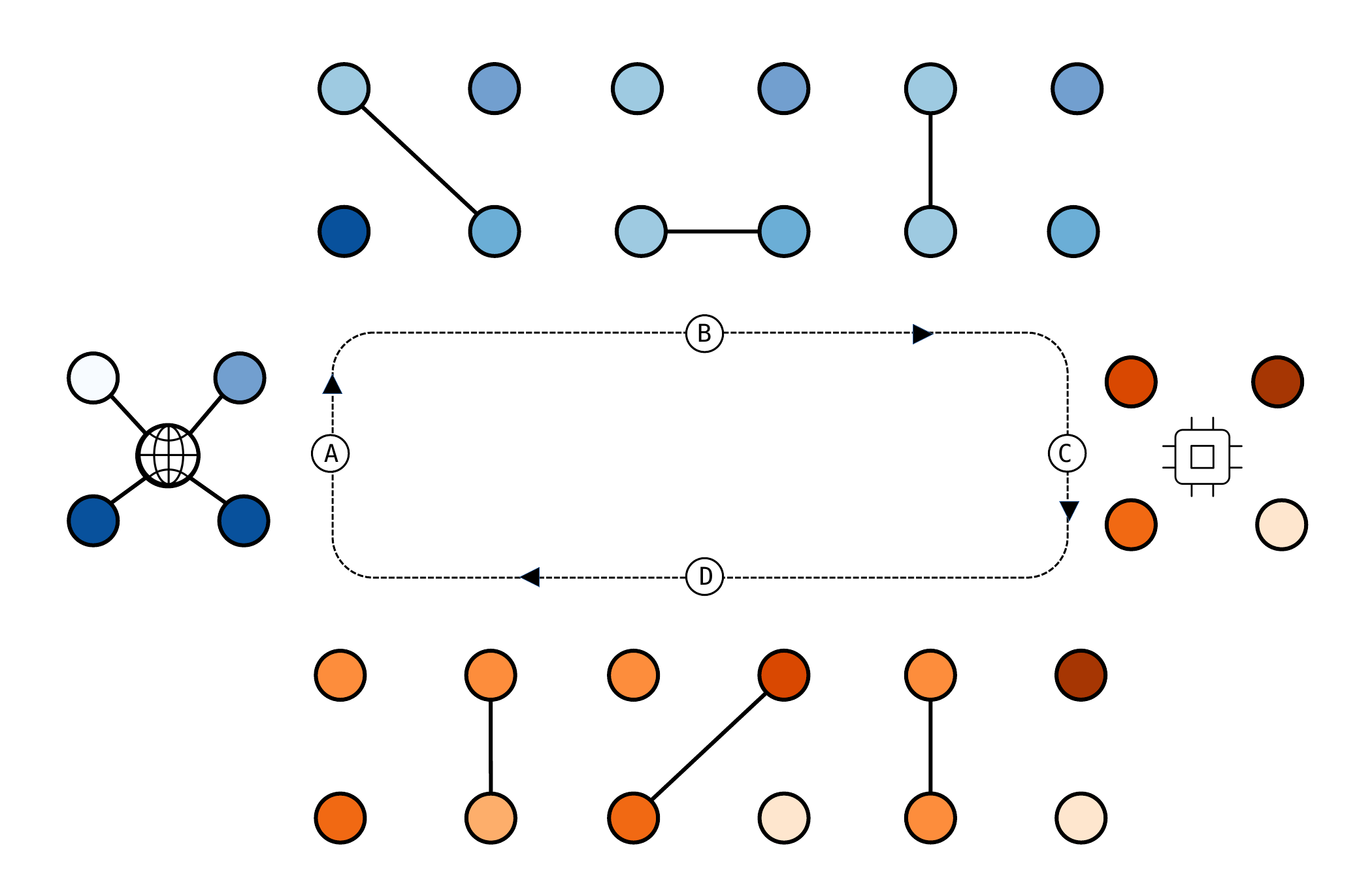}
          \caption{Diagram illustrating the information flow from the algorithm, clockwise from the left.}\label{fig:diagram}
          \Description{Diagram illustrating the information flow for the algorithm.}
    \vspace*{-2ex}   
\end{figure} 

\begin{equation}\label{eqn:Q_consensus}
   y^i_t(k+1) =  \sum_{j\in\Nn^i_k} W^{i, j}_k \cdot y^j_t(k) \quad\forall t\in\Tt^i\text{,} 
\end{equation}
which is performed $K$ times. We thus assign the result from the consensus steps $y^i_t(K+1)$ to the variable $\bar{y}^i_t$. The third step is the parameter update:
\begin{equation}\label{eqn:Q_update}
    \loss{\theta^i; \Tt^i, \bar{y}^i} = \frac{1}{|\Tt^i|}\sum_{\tau^i\in\Tt^i}\left(\bar{y}^i_t - Q(o^i_t,a^i_t;\theta)\right)^2.
\end{equation}
 Finally, the fourth step consists in parameter consensus:
\begin{equation*}
   \theta^i(k+1) =  \sum_{j\in\Nn^i_k} W^{i, j}_k \cdot \theta^j(k)\text{,} 
\end{equation*}
for $K$ rounds. As a result, agents obtain a local approximation of a common average $\theta$. However, for a finite number of consensus steps $K$, it is impossible to guarantee that agents will obtain identical copies $\theta^1=\dots=\theta^N=\theta$. Hence,  agents are left with the parameters from this finite step approximation. 

\section{Experiments}\label{sec:experiments}

We evaluate the performance of DNA-MARL following the methodology outlined by~\citet{papoudakis_2021} for benchmarking multi-agent deep reinforcement learning algorithms in cooperative tasks. This section presents the scenarios, baselines, and evaluation metrics.

\subsection{Scenarios}

Multi-agent environments are typically designed for the cooperative setting~\cite{oroojlooy_2022}, but they can also be configured for the mixed setting. In the mixed setting, individual rewards are emitted, and the team reward is obtained by averaging all individual rewards. With minor adaptations which we outline briefly, the multi-agent particle environment (MPE)~\cite{lowe_2017} scenarios were adjusted for partial observability and individual rewards. The scenarios include:

\textbf{Adversary}\footnote{\label{note:MPE}\texttt{\url{https://github.com/GAIPS/multiagent-particle-envs}}.}: The first MPE adaptation has two teammates protecting a target landmark from a third adversary agent. Teammates are rewarded the adversary's distance from the target and penalized with their negative distance to the landmark. The teammates observations include the position and color from the closest agent (either adversary or teammate), their relative distance to landmark, and the position of the two landmarks. 

\textbf{Spread}\footref{note:MPE}: The second MPE adaptation has three agents that must navigate to three landmarks while incurring a penalty for collisions. We adapt the observation and reward for the partially observable and decentralized setting. Each agent's observation contains its own absolute location, the relative locations of the nearest agent, and the relative location of the nearest landmark. The reward is the negative distance of the agent to the closest landmark.

\textbf{Tag}\footref{note:MPE}: The third MPE adaptation has three big predators (agents) that rewarded for catching a smaller and faster fleeing agent that follows a pre-trained policy. Additionally, two landmarks are placed as obstacles. Agents navigate a two-dimensional grid with continuous coordinates. The reward is sparse, and we adapt the environment for partial observability and decentralization. Each agent's observation includes its own position and velocity, the closest predator's position, and the prey's position and velocity. The reward is individual where the agent that catches the prey is the one receiving a reward of ten points.

\textbf{Level-Based Foraging} (LBF)~\cite{papoudakis_2021}\footnote{\url{https://github.com/semitable/lb-foraging}.}: In this scenario, agents can move on a two-dimensional discrete position grid and collect fruits. Since both agents and fruits have associated levels, successful fruit loading occurs only if the total level of the agents attempting to load it exceeds the fruit's level. Observations consist of relative positions of agents and fruits within a two-block radius centered around the agent. The rewards are sparse, and only the agents that successfully load a fruit receive positive reward. We configure three instances in the partially observable setting, in increasing levels of difficulty: (i) Easy: 10 x 10 grid, 3 players, and 3 fruits (ii) Medium: 15 x 15 grid, 4 players, and 5 fruits (iii) Hard: 15 x 15 grid, 3 players, and 5 fruits.

\begin{figure*}[t]
    \centering
    \begin{center}
        \begin{subfigure}[b]{0.24\textwidth}
            \includegraphics[width=\textwidth]{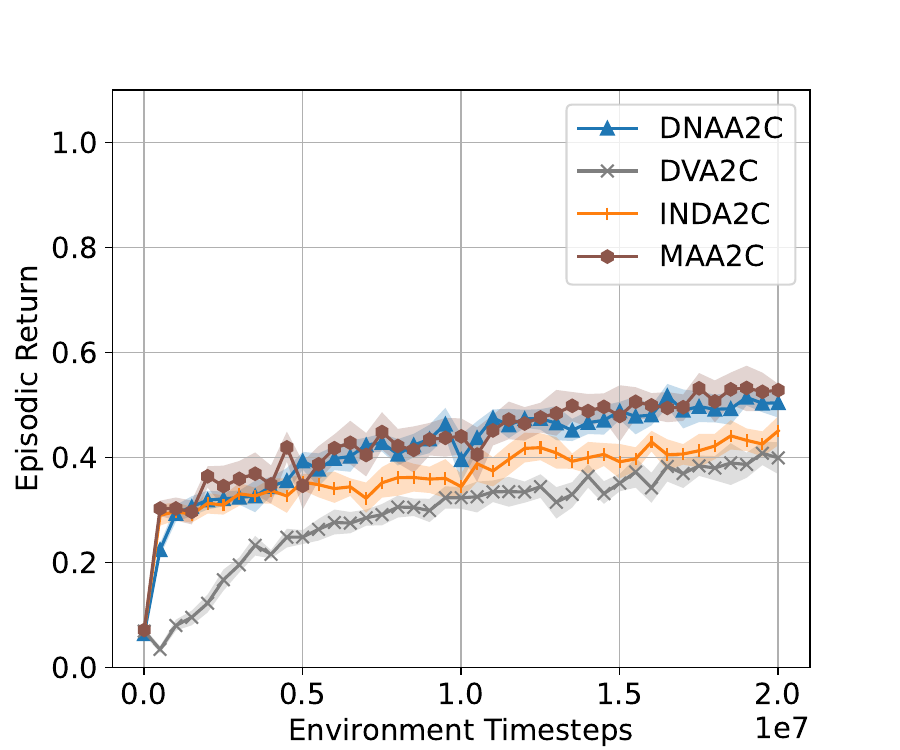}
            \caption{(on-policy) LBF Hard}
        \end{subfigure}
        \begin{subfigure}[b]{0.24\textwidth}
            \includegraphics[width=\textwidth]{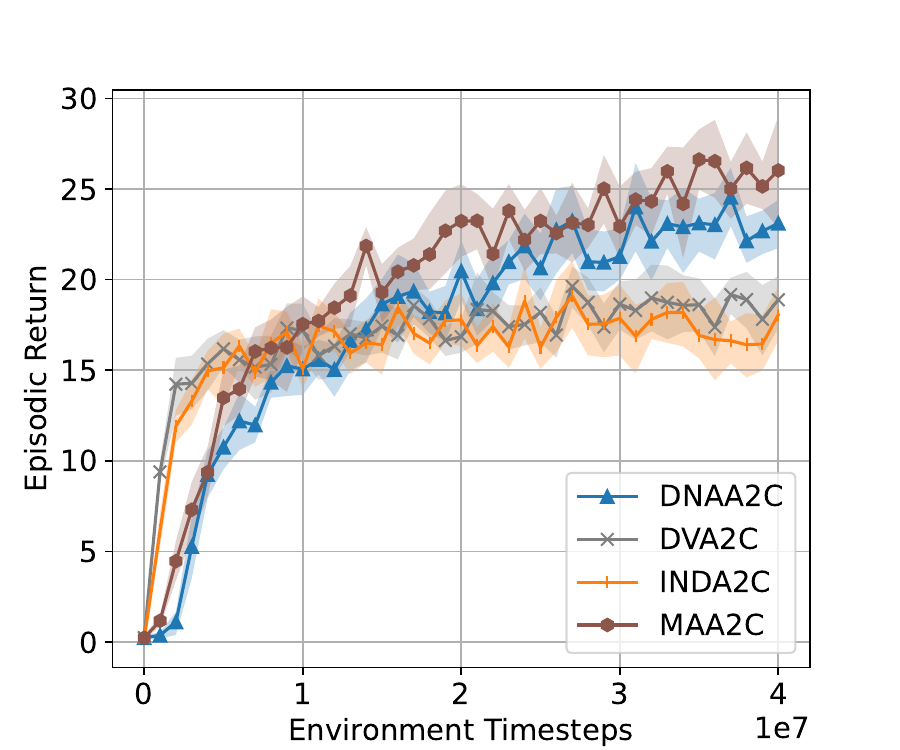}
            \caption{(on-policy) MPE Tag}
        \end{subfigure}
        \begin{subfigure}[b]{0.24\textwidth}
            \includegraphics[width=\textwidth]{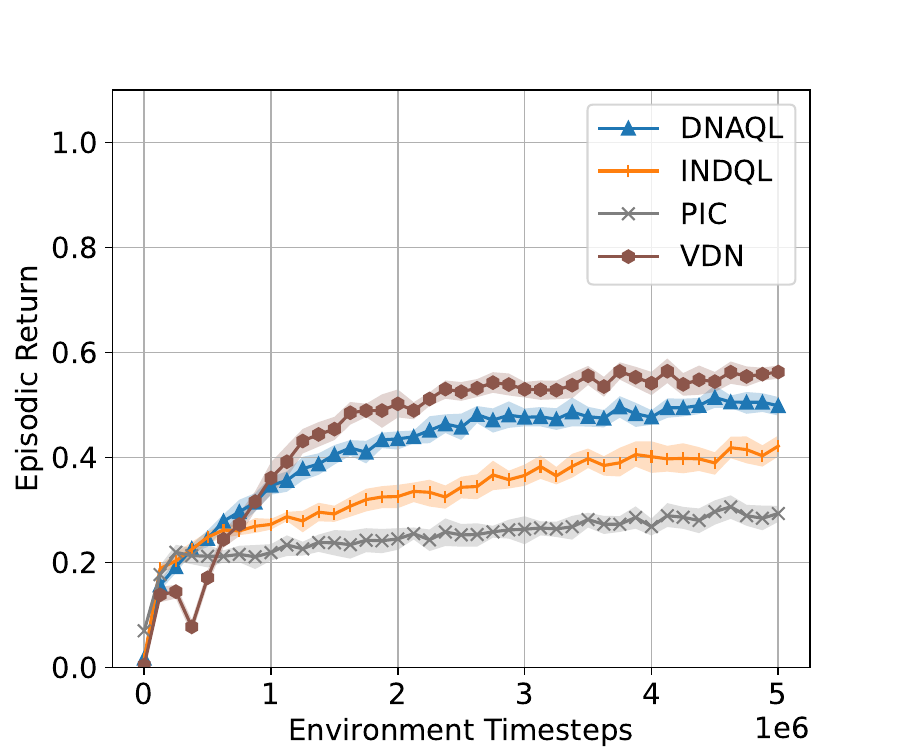}
            \caption{(off-policy) LBF Hard}
        \end{subfigure}
        \begin{subfigure}[b]{0.24\textwidth}
            \includegraphics[width=\textwidth]{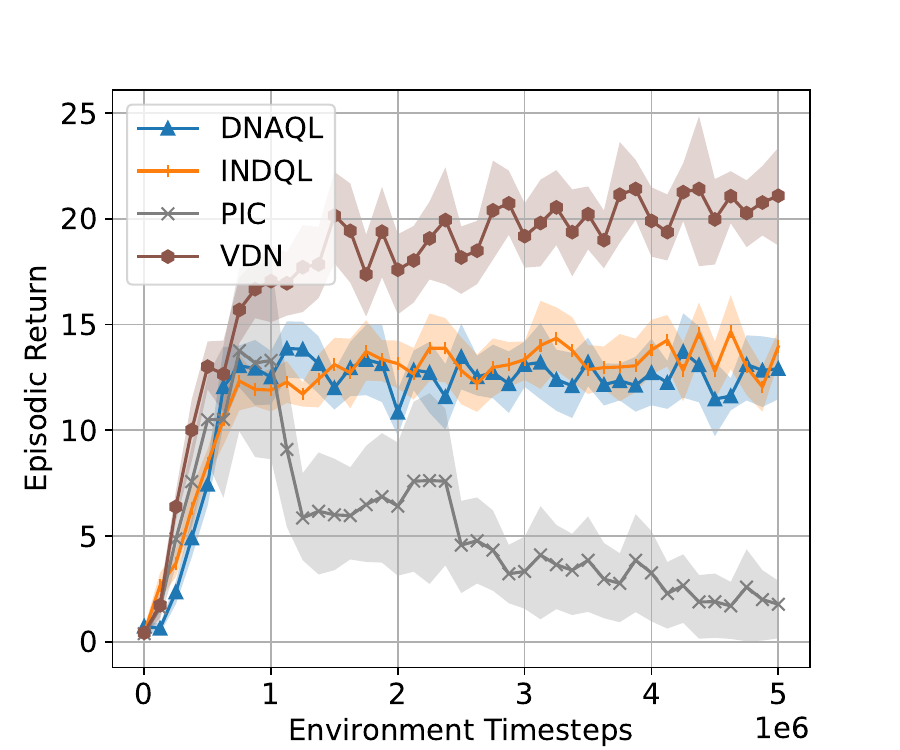}
            \caption{(off-policy) MPE Tag}
        \end{subfigure}
        \Description{Four plots depicting combinations of on-policy and off-policy settings in tasks from the level-based foraging and multi-particle environments.}
    \end{center}
    \caption{From left to right, episodic returns for on-policy setting and episodic returns for the off-policy setting, for two selected tasks. We plot the 95\% bootstrap CI for each algorithm. In chestnut, CTDE algorithms that establish the upper bound of performance. In grey, DTDE algorithms that are DNA-MARL's closest competitors. In orange, IL algorithms that establish a lower bound on performance. We can see that for three algorithms-environments combinations, (a), (b), (c), DNA-MARL (in blue) has the closest performance to the upper bound.}\label{fig:max}
\end{figure*}

\subsection{Baselines}

Following~\citet{papoudakis_2021}, we divide our experiments into the {\em on-policy} and {\em off-policy} settings. Furthermore, for each scenario, we compare three different approaches: (i) individual learners (IL), (ii) decentralized training and fully decentralized execution (DTDE) and (iii) the centralized training and decentralized execution (CTDE) algorithms. ILs are always self interested and CTDE are always fully cooperative as they have access to the team reward.  For the on-policy setting, the baselines include:

\textbf{Multi-Agent Actor-Critic with Advantage} (MAA2C)~\cite{papoudakis_2021}: This is a CTDE algorithm with a central critic that has more information during training: (i) it has access to the {\em joint reward}, (ii) the central critic has access to the concatenation of the all agents' observation, and (iii) uses parameter sharing. Hence, it can compute the {\em team advantage under partial observability} in \eqref{eqn:team_A} precisely.

\textbf{Distributed-V with Advantage} (DVA2C): This DTDE algorithm implements networked agents in Algorithm 2, of~\citet{zhang_2018}, where it performs a consensus on the critic's  network parameters. Moreover, it is a model-based algorithm, whereby it has a neural network that estimates the discounted team return.  

\textbf{Independent Actor-Critic with Advantage} (INDA2C): This IL algorithm implements updates in \eqref{eqn:actor_update} and \eqref{eqn:critic_update} and generates self-interested agents. 

In addition, for the off-policy setting, we specifically use the following baselines:

\textbf{Value Decomposition Networks} (VDN)~\cite{sunehag_2018}: A CTDE algorithm that learns a central $Q$-function that can be factorized among agents.

\textbf{Permutation Invariant Critic} (PIC)~\cite{liu_2020}: Since the DTDE implementation of~\citet{chen_2022} is not open-sourced, we represent its implementation using  the CTDE algorithm in which it was based. PIC has a central critic that employs a graph convolution neural network~\cite{kipf_2017} that learns from joint observations, joint actions and joint rewards. Resulting in $Q$-function representations that remain unchanged regardless of the ordering of the concatenation of observations and actions from the agents. Since PIC sets an upper bound in performance for the networked agents proposed by~\citet{chen_2022} the comparison is fair.

\textbf{Independent Q-Learner} (INDQL): This IL algorithm implements deep $Q$-network updates~\cite{mnih_2015} and generates self-interested agents.

\subsection{Evaluation}\label{sec:evaluation}

In this section we establish the performance metric, the deviation metric and the hypothesis test to discriminate results.

\textbf{Performance metrics}: We evaluate the performance of the algorithms using the maximum average episodic returns~\cite{papoudakis_2021} criteria. For each algorithm, we perform forty one evaluation checkpoints during training each comprising of a hundred rollouts. Then we identify the evaluation time step at which the algorithm achieves the highest average evaluation returns across ten random seeds. Moreover, for each algorithm-scenario combination we report the 95\% bootstrap confidence interval (CI) constructed by resampling the empirical maximum average episodic returns ten thousand times. 

\textbf{Bootstrap hypothesis test}~\cite{colas_2019}: In addition to reporting 95\% bootstrap confidence interval, we gauge how similar two results are by evaluating a bootstrap hypothesis test\footnote{\url{https://github.com/flowersteam/rl_stats}}. The test's null hypothesis is that the means of both samples are the same. The test is performed by drawing an observation from each sample and computing their difference. This procedure is repeated a thousand times. Finally, from the resulting sample of differences we perform the 95\% bootstrap confidence interval. If the CI doesn't contain zero, than we must reject the null hypothesis that both means are equal. We refer to Appendix~\ref{appendix:experiments} for a detailed description of the experimental methodology, hyperparameters, and supplementary results.

\section{Results}

Figure \ref{fig:max} presents a comparative analysis of DNA-MARL's performance for the on-policy and off-policy settings for two selected tasks from the LBF and MPE environments. The CTDE algorithms serve as an upper benchmark for other methods, while the IL algorithms establish a lower performance boundary. Notably, in three specific algorithm-task pairings – (a), (b), and (c) – DNA-MARL demonstrates superior results compared to its nearest competitors when utilizing decentralized training combined with decentralized execution strategies. These results indicate that our double networked averaging A2C (DNAA2C), an algorithm that learns using local observations, can indeed emulate a central critic that uses system-wide observations. In spite of having information loss due to randomized communication. For the off-policy setting, in the MPE environment for the Tag task, our double networked averaging $Q$-learner (DNAQL) has the performance comparable to the IL algorithm, while the alternative decentralized training algorithm has the worst performance. To improve the analysis of these results, we present the maximum average episodic return obtained per algorithm-scenario combination in Table~\ref{tab:max}.

\begin{table*}
   \setlength{\tabcolsep}{.16667em}
    \begin{center}
    \caption{Results for level-based foraging and multiagent particle environments: Maximum average episodic returns over ten independent runs and 95\% bootstrap confidence interval. Highlighted results indicate the best performing algorithm. The asterisk indicates results that are not significantly different from the best result. Double asterisks indicate the second best result.}\label{tab:max}
    {\tabulinesep=0.3mm
        \begin{tabu}{llccccccc}
            \toprule
                                    & & \multicolumn{3}{c}{\textbf{LBF}} & &\multicolumn{3}{c}{\textbf{MPE}}\\\cline{3-5}\cline{7-9}
               \textbf{Methods} & \textbf{Algorithm} &    \textbf{Easy} & \textbf{Medium} & \textbf{Hard} &   & \textbf{Adv.} & \textbf{Spread} & \textbf{Tag}\\
             \midrule
             \parbox[t]{2mm}{\multirow{4}{*}{\rotatebox[origin=c]{90}{on-policy}}} & MAA2C (CTDE) & \CB{0.96}{-0.01, 0.01} & \CB{0.76}{-0.04, 0.04} & \CB{0.53}{-0.04, 0.04} &   & \CI{17.39^*}{-0.56, 0.62} & \CI{-92.19^*}{-0.35, 0.36} & \CB{26.63}{-1.68, 1.70}\\ \cline{2-9}
                    & DNAA2C (ours) &  \CI{0.93^{**}}{-0.01, 0.01}  & \CI{0.75^*}{-0.02, 0.02} & \CI{0.52^*}{-0.02, 0.02} &  &\CB{17.68}{-0.67,0.68} & \CB{-91.96}{-0.28, 0.24} & \CI{26.09^*}{-2.11, 2.01}\\
                    & DVA2C &  \CI{0.83}{-0.01, 0.01} & \CI{0.67}{-0.04, 0.04} &  \CI{0.41}{-0.02, 0.02} &   & \CI{16.46}{-0.89, 0.70} & \CI{-93.49}{-0.78, 0.78} & \CI{19.63}{-1.17, 1.07} \\
                    & INDA2C (IL) &  \CI{0.89}{-0.02, 0.01} &  \CI{0.69}{-0.03, 0.03} & \CI{0.45}{-0.03, 0.03}  &  & \CI{16.30}{-0.68, 0.67} & \CI{-94.39}{-0.52, 0.52} & \CI{19.10}{-1.74, 1.90}\\\hline
             \parbox[t]{2mm}{\multirow{4}{*}{\rotatebox[origin=c]{90}{off-policy}}} & VDN (CTDE) &  \CB{0.94}{-0.01, 0.01} &  \CB{0.79}{-0.02, 0.02} & \CB{0.56}{-0.02, 0.02} & & \CI{9.64}{-0.64, 0.77} & \CI{-94.77}{-0.26, 0.28} & \CB{23.27}{-2.69, 2.69}\\ \cline{2-9}
                    & DNAQL (ours) & \CI{0.88^{**}}{-0.01, 0.01} & \CI{0.75^{**}}{-0.02, 0.02} & \CI{0.51^{**}}{-0.02, 0.02} & & \CI{12.51^*}{-1.05, 0.92} & \CI{-93.06^*}{-0.45, 0.48} & \CI{15.77}{-1.70, 1.88}\\
                    & PIC &  \CI{0.48}{-0.03, 0.04} &   \CI{0.48}{-0.02, 0.02} &  \CI{0.31}{-0.02, 0.02} & & \CI{11.10}{-0.96, 0.92} & \CI{-93.94}{-0.41, 0.39} & \CI{13.76}{-3.72, 3.94}\\
                    & INDQL (IL) &  \CI{0.86}{-0.02, 0.01} &  \CI{0.70}{-0.01, 0.01}  & \CI{0.42}{-0.02, 0.02} & & \CB{13.61}{-1.13, 1.02} & \CB{-92.69 }{-0.33, 0.32} & \CI{15.54}{-1.10, 1.27}\\\hline
            \bottomrule
        \end{tabu}}
    \end{center}
\end{table*} 

In Table~\ref{tab:max}, results are separated into on-policy and off-policy settings, and the CTDE algorithms is outlined for each method. For each algorithm-scenario pairings, the values highlighted in bold represent the best results for the scenario. The asterisk shows the results of the bootstrap hypothesis test cannot reject the null hypothesis (\ie, the performance is equal to the best performing algorithm for the scenario). The double asterisk indicates results that are worse than the highlighted result but still outperform the other algorithms, according to the bootstrap hypothesis test.

For the on-policy setting, we note, from six scenarios, there are four scenarios that the results are comparable to MAA2C's performance. In addition, in the remaining two scenarios, both results are very close. For the off-policy setting and LBF environment we observe similar outcomes: DNAQL ranks as second best. Moving to MPE environment for the off-policy setting, the IL method outperforms the CTDE for two scenarios, followed by DNAQL. This is not a surprising result, since~\citet{papoudakis_2021} state that for most MPE scenarios, VDN's assumption of additive value function decomposition is mostly violated for this environment. However, for the third scenario, Tag, VDN outperforms other decentralized algorithms by a wide margin. Here, additive value decomposition seems to have played a major role in the performance. Additive value decomposition limits the range of representable functions, but simplifies the learning over a large combined observation and actions spaces. Neither DNAQL or INDQL are guaranteed to generate additive value decomposition. 

\begin{table*}[t]
   \setlength{\tabcolsep}{.16667em}
    \begin{center}
    \caption{MARL settings. The codes for reward column: Individual (I), or team (T). The codes for state space observability (Observ.): Fully observable (FO), joint fully observable (JFO), and partially observable (PO). The codes for training (Train.) column: Centralized (C), or decentralized (D). The codes for the base Markov decision problem framework~\cite{oliehoek_2016}: Markov game (MG), decentralized Partially observable Markov decision process (dec-POMDP), homogenous Markov game (HMG), and partially observable Markov game (POMG). The is homogeneous column requires a special structure on agents.}\label{tab:works}
    {\tabulinesep=0.3mm
        \begin{tabu}{lcccccc}
            \toprule
                \tabbreak{}{Works} &  \tabbreak{}{Reward} & \tabbreak{}{Observ.} & \tabbreak{}{Train.} &  \tabbreak{Base}{Framework} & \tabbreak{Communicates}{Observ.} & \tabbreak{Is}{Homogeneous}\\
             \midrule
              \citet{lowe_2017}      &  T/I & PO & C &  Dec-POMDP/POMG & No & Heterogeneous\\
              \citet{sunehag_2018}   &  T & PO & C &  Dec-POMDP & No & Heterogeneous\\    
              \citet{zhang_2018}     &  I & FO & D &  MG & No & Heterogeneous\\     
              \citet{chen_2022}      &  I & JFO & D &  HMG &  Yes & Homogeneous\\     
              DNA-MARL                &  I & PO & D &  POMG & No & Homogeneous\\
            \bottomrule
        \end{tabu}}
    \end{center}
\end{table*} 

\balance

\subsection{Ablations}
To assess the impact of consensus steps within the DNA-MARL framework, we conducted ablations regarding the discounted return consensus outlined in \eqref{eqn:V_consensus}, specifically focusing on both the critic parameters and actor parameters.

\begin{figure}[t]
    \centering
    \begin{center}
        \begin{subfigure}[b]{0.4\linewidth}
            \includegraphics[width=\linewidth]{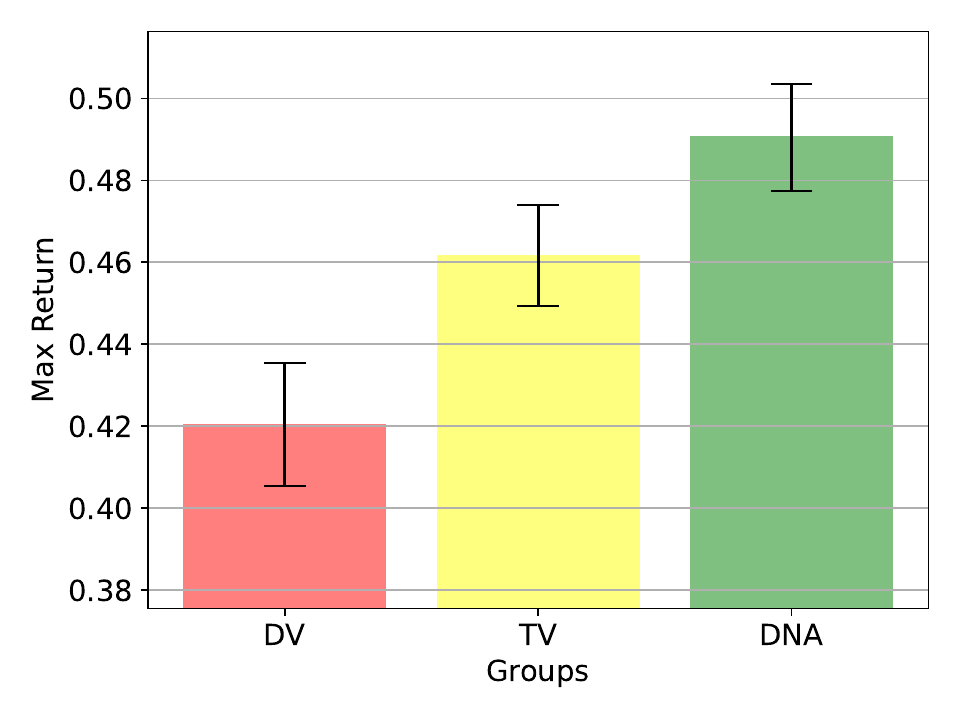}
            \caption{LBF Hard}
        \end{subfigure}
        \begin{subfigure}[b]{0.4\linewidth}
            \includegraphics[width=\linewidth]{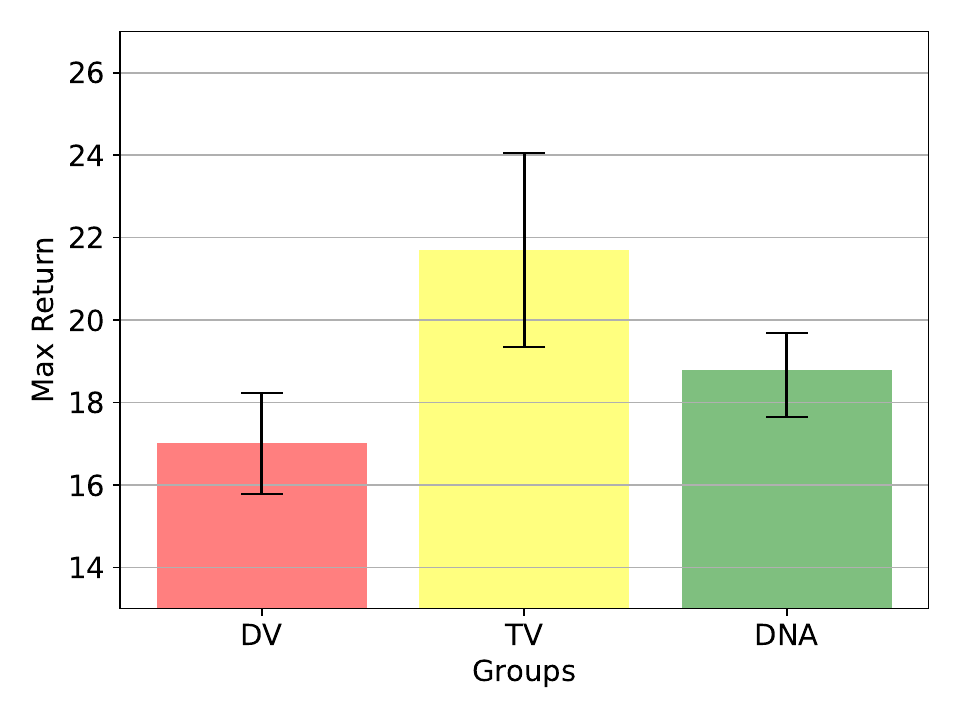}
            \caption{MPE Tag}
        \end{subfigure}
\end{center}
  \caption{Ablation for DNAA2C: From left to right, DV (distributed-V) group has critic consensus. TV (team-V) group has team-$V$ consensus and critic consensus. DNA group has team-$V$ consensus and both actor and critic consensus. We can see a performance improvement moving from DV to TV which highlights the impact of our contribution.}\label{fig:ablation}
    \Description{Ablation plots for on-policy setting for environments level-based foraging and multi-particle environment.}
\end{figure}

Figure~\ref{fig:ablation} presents the ablation results for DNAA2C in the tasks LBF Hard and MPE Tag. The bars represent the averages within a neighborhood of size two around the maximum average episodic returns, while the intervals denote the 95\% bootstrap confidence interval. Moving from left to right, we have three groups: (i) Distributed-V (DV) which conducts consensus on the critic parameters, (ii) Team-V (TV) which performs consensus on both the $V$-values and critic parameters, and (iii) DNA which carries out consensus on both the $V$-values and actor-critic parameters. 
Due to space restrictions the remaining ablation plots are reported in (Appendix~\ref{appendix:results}). Here, we highlight how this ablation study connects DNA-MARL to previous works:~\citet{zhang_2018} propose critic parameter consensus (DV group). Our original contribution proposes consensus on the team-$V$ \eqref{eqn:V_consensus} in addition to consensus update on the critic's parameters (TV group): While Fig.~\ref{fig:ablation} (a) shows that team-$V$ provides an improvement in overall performance for LBF Hard task, Fig.~\ref{fig:ablation} (b) shows that team-$V$ consensus provides a significant improvement in performance for the MPE Tag task. Finally, DNA-MARL combines team-$V$ consensus with consensus on the agents' policies proposed by~\citet{chen_2022} emulating {\em parameter sharing} in the decentralized setting. 

\section{Related Work}

We relate our work with five other lines of research, two of which we present herein: the centralized training and decentralized execution under partial observability setting and networked agents in the decentralized training and fully decentralized execution (DTDE). Due to space restrictions, we further discuss related works in Appendix~\ref{appendix:related_work}.

\textbf{Central Training and Decentralized Execution}: CTDE is the prevailing approach in multi-agent reinforcement learning, where a \textit{central critic} learns a system action-value function to mitigate the risk of non-stationarity. The policies are factorized and executed by individual actors, utilizing only local information to address the large state space problem. Examples from works that learn a central critic include MADDPG~\cite{lowe_2017}, COMA~\cite{foerster_2018} and PIC~\cite{liu_2020}. Furthermore, actors benefit from {\em parameter sharing} as proposed by~\cite{gupta_2017}, wherein agents use a single neural network to approximate a policy trained with experiences collected from the behavior policies of all agents. Parameter sharing reduces wall clock time and increases sample-efficiency, enabling faster agent learning~\cite{hernandez-leal_2019}. Another possibility is building utility functions that factorize into agent-wise function.~\citet{sunehag_2018} propose value decomposition networks, where the team-$Q$ function is recovered by adding the agent-wise $Q$-functions. Finally, QMIX~\cite{rashid_2018} extend VDN by proposing a mixing network, that has a dynamic set of parameters that vary according to the system state. The mixing network produces more expressive team-$Q$ function decomposition, requiring that the joint action that minimizes the team $Q$-value be the same as the combination of the individual actions maximizing the agent-wise $Q$-values.

\textit{Networked agents with multi-agent reinforcement learning}.~\citet{zhang_2018} is DTDE MARL system that apply the consensus mechanism over the critic's parameters to obtain a joint policy evaluation. However, their system requires full observability of both state and action spaces. In contrast~\citet{zhang_2019} propose DTDE MARL system that performs consensus on the actor's parameters while the critics are individual. As a limitation the policies must represent the joint action space. ~\citet{chen_2022} apply networked agents to homogeneous Markov games, a subclass of Markov game, where agents observe different permutations of the state space but share the same action space, making individual agents interchangeable. To improve observability agents choose when and to whom communicate their observation. Differently from other approaches our agents obtain team-$V$ estimation using consensus. Experimental results indicate that DNA-MARL outperforms both~\cite{zhang_2018} and~\cite{chen_2022} under partially observable settings. Table~\ref{tab:works} summarizes the differences between our DNA-MARL method and previous networked agents systems.

\section{Conclusion and Future Work}

We propose the DNA-MARL that learn to cooperate in a ND-POMG under the decentralized training and fully decentralized execution paradigm. The key is  performing consensus steps on the $V$-values. Our experiments show that DNA-MARL agents, with limited access to system information, can often reach the performance of their centralized training counter parts and outperform previous works. Moreover, the framework is quite generic, offering opportunities for extensions of popular single agent algorithms, \eg, TRPO~\cite{schulman_2015}, PPO~\cite{schulman_2017}. And also combine them with multi-agent belief systems.



\begin{acks}
This work was supported by national funds through Fundação para a Ciência e a Tecnologia (FCT) with reference UIDB/50021/2020 - DOI: 10.54499/UIDB/50021/2020. Center for Responsible AI - C628696807-00454142. Guilherme S. Varela is supported by FCT scholarship 2021.05435.BD.
\end{acks}



\bibliographystyle{ACM-Reference-Format} 
\bibliography{sample}


\onecolumn  
\appendix
\pagenumbering{roman}
\section{Extended Background}\label{appendix:metropolis}

Consider a graph $\Gg(\Nn, \Ee)$, an undirected graph with nodes $\Nn = \{1, \ldots, N\}$ and
edges $\Ee\subseteq \Nn\times\Nn$. An edge in $\Gg$ is represented by $(m, n) \in \Ee$, where $m, n \in \Nn$. The set of neighbors of a node $n$ is denoted by $\Nn_n = \{ m \mid (n, m) \in \Ee\}$, and the degree of node $n$ is given by its cardinality: $d(n) = |\Nn_n|$.

Let $\Gg(\Nn, \Ee)$ represent an arbitrary connected graph with nodes $\Nn = \{1, \ldots, N\}$ and edges $\Ee\subseteq \Nn\times\Nn$. The Metropolis weights are defined by a matrix $W \in \RR^N\times \RR^N$, where:
\begin{equation}\label{eqn:metropolis_hastings}
      W(n, m) =
       \begin{cases}
           1 - \sum_{m'\neq n} W(n, m') & \quad \text{if } n=m \\
           \frac{1}{1 + \max\left(d(n), d(m)\right)} & \quad \text{if $(n,m) \in \Ee$} \\
           0 & \quad \text{otherwise}
       \end{cases}
 \end{equation}
As $\Gg$ is connected—meaning there exists a path between any two nodes—the Metropolis weights matrix asymptotically ensures the average consensus, as demonstrated by~\citet{xiao_2007}. Here, $d(n)$ and $d(m)$ denote the degrees of nodes $n\in \Nn$ and $m\in \Nn$, respectively.

 For instance, let's consider a connected graph $G$, represented by its adjacency matrix $A_G=\left[a_{n,m}\right]$\footnote{By definition $A_G$ is such that the element $a_{n,m}=1$ if $n,m\in \Ee$ and $a_{n,m}=0$ otherwise.}.  The Metropolis weights $W_G$ for $A_G$ is given by: 
 \begin{equation*}
        A_G=
       \begin{bmatrix}
        0 & 0 & 1 & 1 & 1\\
        0 & 0 & 1 & 0 & 1\\
        1 & 1 & 0 & 1 & 1\\
        1 & 0 & 1 & 1 & 1\\
        1 & 1 & 1 & 1 & 0
       \end{bmatrix}
       \quad\text{then}\quad
        C_G=
       \begin{bmatrix}
        0.35& 0 & 0.2 & 0.25 & 0.2\\
        0 & 0.6 & 0.2 & 0 & 0.2\\
        0.2 & 0.2 & 0.2 & 0.2 & 0.2\\
        0.25 & 0 & 0.2 & 0.35 & 0.2\\
        0.2 & 0.2 & 0.2 & 0.2 & 0.2
       \end{bmatrix}\text{,}
 \end{equation*}
according to \eqref{eqn:metropolis_hastings}.  The Metropolis weights can be determined locally by performing two rounds of communication. In this processe, each node probes the network for neighbors and then  communicates its degree to them. Each node computes $max(d(n), d(m))$ locally, and parameter communication may resume. It should be noted that no node has knowledge of matrix $\Gg$ or even the total number of peers $N$~\cite{xiao_2007}.

\section{Extended Double Networked Averaging}\label{appendix:DNAMARL}
 
This section is divided into two parts. The first part formalizes the {\em graph model}. The second part provides the pseudocode for the double networked agents.

\subsection{Graph Model}\label{appendix:graph_model}

Let $\Gg_k$ represent a switching topology communication network, wherein nodes correspond to reinforcement learning agents and the edges represent the communication links between agents. The {\em dynamics} of the network is described by its {\em graph model}. In this work,  the graph model $\Gg_k=(\Nn, \Ee_k)$  characterizes  an undirected graph where $\Nn$ is the fixed set of $N$-agents, and $\Ee_k$ is the edge-set representing the communication links between agents at time $k$. We consider a random sequence of edge sets $\Ee_k$ drawn from  a finite {\em collection of edge sets} $\Ee$:

\begin{equation}\label{eqn:edge_sets}\{\Ee_k\}_{k\ge  0}\in \Ee = \{E_1, ..., E_\text{J}\}\text{,} \end{equation}
such that every edge set $E_j\in \Ee$ has a positive probability  of composing the graph $\Gg_k$: 
\begin{equation*}\PP[]{\Ee_k=E_j}> 0\quad \wedge \quad  \sum^\text{J}_{j=0}\PP[]{\Ee_k=E_j} = 1\text{.}\end{equation*}
The edge sets are independently drawn from $\Ee$ , \ie, $\Ee_k$ is independent of $\Ee_s$ for $k\neq s$. We call a finite collection of graphs with a common node set $\Nn$ and edge sets $\Ee_k, k=1,\dots, p$ {\em jointly connected} if 
\begin{equation*}\Gg(\Nn, \bigcup^p_{k=0} E_k)\quad\text{is connected}\text{.}\end{equation*}
The collection of communication graphs induced by the switching topology dynamics must be jointly connected for agents to reach consensus  (Theorem 1,~\citet{xiao_2007}).  As a consequence, we build the collection of edge sets $\Ee$ in \eqref{eqn:edge_sets} considering all the possible edge sets for fixed number of edges. And we define the distribution $\beta$ as an uniform distribution over $\Ee$, \ie, 
\begin{equation*}\forall E_j\in\Ee\quad \PP[]{\Ee_k=E_j}=\frac{1}{\text{J}}\text{.}\end{equation*}

 \subsection{Pseudocodes}\label{appendix:pseudocodes}

 Listing~\ref{alg:DNAA2C} reports the serial implementation of a synchronous distributed algorithm DNAA2C. The artifact consists of four main blocks: (i) In lines 4-11, agents interact with environment. (ii) In lines 12-22, agents compute team-$V$. (iii) In lines 23-26, agents perform local mini-batch gradient descent, for actor and critic error minimization , and (iv) in lines 27-36, agents perform actor-critic consensus. Although the implementation is serial, it can be run in parallel. Assuming that agents can interact with the environment independently but at the same time slots.  Agents perform the local updates in parallel.
 
 Specifically, the initialization (lines 1-2) involves $N$ independent pairs of parameters: $\theta^i$ for each agent's actor module and $\omega^i$ for each critic module. The following hyperparameters control various aspects of the training process:

 \begin{itemize}
     \item $T_{max}$ maximum number of episodes.
     \item $K$ number of consensus rounds.
     \item $I$ the training interval between two rounds of consensus on the parameters.
     \item $C$ the number of edges on the communication graph.
 \end{itemize}

 Additionally, we set the critic's parameters to the same value (the average of all $\omega^i$) and initialize a memory for the trajectories $\tau$.

 \begin{algorithm}[H]
     \caption{Double networked actor-critic with advantage}\label{alg:DNAA2C}
     \begin{algorithmic}[1]
        \REQUIRE ${\omega^i}_{i\in\Nn}$, ${\theta^i}_{i\in\Nn}$, $T_{max}$,  $K$, $I$,$C$
        \STATE $\omega^i\gets \bar{\omega}\quad \forall i\in\Nn$
        \STATE $\tau\gets \emptyset$
        \WHILE{$p\leq T_{max}$}
             \STATE $s \sim \mu(\cdot), t\gets 0$  
             \WHILE {$s\neq s_{\text{terminal}}$} 
                 \STATE Observe $o^1_t,\dots, o^N_t$
                 \STATE Sample actions $a^i\sim\pi_{\theta^i}(\cdot|o^i_t)\quad \forall i=1,\dots, N$
                 \STATE Execute actions and collect $r^1_{t+1},\dots, r^N_{t+1}$
                 \STATE $\Tt^i \gets\Tt^i \cup \left\{(o^i_t, a^i_t, r^i_{t+1})\right\}_{i\in\Nn}$
                 \STATE $s\gets s', t\gets t+1$
            \ENDWHILE  
            \FOR{$i\in \Nn$}
                \STATE $y^i_t = r^i_{t+1} + \gamma V(o_{t+1};\omega^i) \quad\forall t\in \Tt^i$ 
                \STATE $\bar{y}^i \gets \{y^i_t\}^T_{t=0}$
            \ENDFOR 
            \FOR{$k=1, \dots K$}
                \STATE Sample edge set $\Ee_k\sim \mathcal{E}(C)$.
                \STATE  Listen to channel $W_k\gets \Gg_k(\Nn, \Ee_k)$.
                \FOR {$i\in\Nn$}
                     \STATE $\bar{y}^i \gets \sum_{j\in\Nn_i} W^{i, j}_k \cdot \bar{y}^j$
                \ENDFOR
            \ENDFOR
            \FOR {$i\in\Nn$} 
                 \STATE Update $\omega^i$ by min. $ \mathcal{L}(\omega^i_{\tau}; \Tt^i, \bar{y}^i)$  Eqn.~\ref{eqn:step_3_team_critic}.
                 \STATE Update $\theta^i$ by min. $\mathcal{L}(\theta^i_{\tau}; \Tt^i, \bar{y}^i)$   Eqn.~\ref{eqn:step_4_team_actor}.
            \ENDFOR
             \IF {$I\,\text{divides}\, p$} 
                 \FOR{$k=1, \dots K$}
                     \STATE Sample edge set $\Ee_k\sim \Ee(C)$.
                     \STATE  Listen to channel $W_k\gets \Gg_k(\Nn, \Ee_k)$.
                     \FOR {$i\in\Nn$}
                         \STATE $\omega^i \gets \sum_{j\in\Nn_i} W^{i, j}_k \cdot \omega^j$
                         \STATE $\theta^i \gets \sum_{j\in\Nn_i} W^{i, j}_k \cdot \theta^j$
                     \ENDFOR
                 \ENDFOR
             \ENDIF
             \STATE $\Tt\gets \emptyset$, $p\gets p+1$
        \ENDWHILE
     \end{algorithmic}
 \end{algorithm}
 
 The training loop spans lines 3-38 and is standard in reinforcement learning~\cite{sutton_2018}. The agents' interaction with the environment spans line 4-11, where initially the environment is set to a hidden state $s$, and the variable $t$ controls the training step (line 5). Agents observe, locally, $o^i_t$ (line 6), and draw actions from their stochastic policies (line 7). The environment emits individual rewards and transitions to the next state $s'$ (line 8). The transition $(o^i_t, a^i_t, r^i_{t+1})$ is appended to the local memory buffer $\tau^i$ (line 9). The environment updates the state to $s'$ and increments the time step counter (line 10). The interaction loop continues until the terminal state is found and $T$ receives $t$.

 The team-$V$ computation sub-routine spans lines 12-22, and is further sub-divided in two blocks: Lines 12-15, agents perform local $V$-value estimation, and lines 16-22 where agents perform consensus updates. From the local trajectory $\tau^i$ agents compute the value for observation $o^i_t$, by using the critic's target network with parameters $\omega^i_-$ \eqref{eqn:local_critic} for all time steps (line 13). We stack all observations on a vector $\bar{y}^i\in \RR^{T}$ (line 14). The team-$V$ consensus loop (line 16) consists of $K$ consensus rounds: The nature draws an edge set $E$ (line 17), the communication graph $\Gg(\Nn, E)$ is determined, making it possible to determine $W_k^{i, j}$ locally (line 18). Lines 19-22 consist of team-$V$ consensus updates in \eqref{eqn:V_consensus}.

 The local mini-batch update loop spans lines 23-27 and, in a synchronous distributed system, it can be performed in parallel. The first local update is the critic update (line 24), that improves the joint policy's evaluation. The second local update is the actor update (line 25) to adjust the local policy in a direction that improves the evaluation for the joint action.

\begin{algorithm}[H]
    \caption{Double networked averaging $Q$-learner}\label{alg:DNAQL}
    \begin{algorithmic}[1]
        \REQUIRE  ${\theta^i}_{i\in\Nn}$, $T_{max}$,  $K$, $I$,$C$
        \STATE $\theta^i\gets \bar{\theta}\quad \forall i\in\Nn$
        \STATE $\tau\gets \emptyset$
        \WHILE{$p\leq T_{max}$}
            \STATE $s \sim \mu(\cdot), t \gets 0$ 
            \WHILE {$s\neq s_{\text{terminal}}$} 
                \STATE Observe $o^1_t,\dots, o^N_t$
                \STATE Sample actions $a^i\sim\pi_{\theta^i}(\cdot|o^i_t)\quad \forall i=1,\dots, N$
                \STATE Execute actions and collect $r^1_{t+1},\dots, r^N_{t+1}$
                \STATE $\Tt^i \gets\Tt^i \cup \left\{(o^i_t, a^i_t, r^i_{t+1})\right\}_{i\in\Nn}$
                \STATE $s\gets s', t\gets t+1$
            \ENDWHILE
            \FOR{$i\in \Nn$}
                \STATE  $y^i_t = r^i_{t+1} + \gamma \max_{u^i_t} Q(o^i_{t+1}, u^i_t;\theta^i_{-})\quad \forall t\in \Tt^i$ 
                \STATE $\bar{y}^i\gets \{y^i_t\}^T_{t=0}$
            \ENDFOR
            \FOR{$k=1, \dots K$}
                \STATE Sample edge set $\Ee_k\sim \Ee(C)$.
                \STATE  Listen to channel $W_k\gets \Gg_k(\Nn, \Ee_k)$.
                \FOR {$i\in\Nn$}
                    \STATE $\bar{y}^i\gets \sum_{j\in\Nn_i} W^{i, j}_k\cdot \bar{y}^j$
                \ENDFOR
            \ENDFOR 
            \FOR {$i\in\Nn$}
                \STATE Update $\theta^i$ by minimizing  \eqref{eqn:Q_update}
            \ENDFOR
            \IF {$I\,\text{ divides }\, p$} 
                \FOR{$k=1, \dots, K$}
                    \STATE Sample edge set $\Ee_k\sim \Ee(C)$.
                    \STATE  Listen to channel $W_k\gets \Gg_k(\Nn, \Ee_k)$.
                    \FOR {$i\in\Nn$}
                        \STATE $\theta^i \gets \sum_{j\in\Nn_i} W^{i, j}_k \cdot \theta^j$
                    \ENDFOR
                \ENDFOR
            \ENDIF
            \STATE $\tau\gets \emptyset, p\gets p + 1$
        \ENDWHILE
    \end{algorithmic}
\end{algorithm}

 Periodically, at every $I$ environment episodes, we perform consensus over the critic and actor parameters to benefit from distributed trajectory collection, resulting in faster sample-efficiency~\cite{chen_2022}. Lines 29-30 are analogous to lines 17-18, where nature draws an edge set from the communication graph. The selected edge set regulates the communication graph's topology. In lines 31-34, agents perform consensus on the critic and actor parameters, noting that the both actor and critic parameters can be pushed to the communication graph at the same time slot.
 
 Listing~\ref{alg:DNAQL} reports the serial implementation of a synchronous distributed algorithm DNAQL. The artifact consists of four main blocks: (i) In lines 4-11, agents interact with the environment. (ii) In lines 12-22, agents compute team-$Q$.  (iii) In lines 23-25, agents compute local mini-batch gradient descent for $Q$-function minimization, and (iv) in lines 26-34, agents perform $\theta$ parameter consensus. Although the implementation is serial, it can be run in parallel. Assuming that agents can interact with the environment independently but at the same time slots. Team $Q$ consensus and parameter consensus can be performed locally. Agents perform the local updates in parallel.
 
Specifically, the initialization (lines 1-2) consists of $N$ independent parameters $\theta^i$ one for each agent. The following hyperparameters control various aspects of the training process:

 \begin{itemize}
     \item $T_{max}$ maximum number of episodes.
     \item $K$ number of consensus rounds.
     \item $I$ the training interval between two rounds of consensus on the parameters.
     \item $C$ the number of edges on the communication graph.
 \end{itemize}
Moreover, we set the parameters $\theta^i$ to the same value and initialize a memory for the trajectories $\tau$.

 The training loop spans lines 3-36 and follows standard reinforcement learning procedures~\cite{sutton_2018}. The agents' interaction with the environment occurs between lines 4-11. Initially, the environment is set to a hidden state $s$, and the variable $t$ controls the training step (line 5). Agents locally observe, their observations $o^i_t$ (line 6) and select actions based on deterministic policies using epsilon greedy criteria~\cite{sutton_2018} (line 7). The environment emits individual rewards and transitions to the next state $s'$ (line 8). The transition $(o^i_t, a^i_t, r^i_{t+1})$ is appended to the local memory buffer $\tau^i$ (line 9). The environment updates the state to $s'$ and increments the time step counter (line 10). The interaction loop continues until the terminal state is reached.
 
 The team-$Q$ computation sub-routine spans lines 12-22, and is divided in two blocks: in lines 12-15, agents perform local $Q$-value estimation, and in lines 16-22, agents execute consensus updates. From the local trajectory $\tau^i$ agents compute $Q$-values for every local action $a^i\in \Aa$ in response to observation $o^i_t$ using the $Q$ target network with parameters $\theta^i_-$ \eqref{eqn:Q_local}, for all time steps (line 13). These $Q$-values are stacked into the vector $\bar{y}^i\in \RR^T$ (line 14). 
 
 The team-$Q$ consensus loop (line 16) consists of $K$ consensus rounds: nature draws an edge set $\Ee_k$ (line 17), and the communication graph $\Gg_k(\Nn, \Ee_k)$ is instantiated, making it possible to determine $W_k^{i, j}$ locally (line 18). Lines 19-22 consist of team-$Q$ consensus updates in \eqref{eqn:Q_consensus}.
 
 The local mini-batch update loop spans lines 23-26 and, in a synchronous distributed system, can be performed in parallel. The local gradient update is given in the direction of the action that maximizes average of $Q$ values (line 24).
 
 Periodically, at every $I$ environment episodes, we perform consensus on the $Q$-network's parameters to benefit from distributed trajectory collection, resulting in faster sample-efficiency~\cite{chen_2022}. Lines 28-29 are analogous to lines 17-18 where nature draws an edge set from the communication graph. The selected edge set regulates the communication graph's topology. In line 31, agents perform consensus on the $Q$-network's parameters, finishing the iteration.
 
 
\section{Extended Experiments}\label{appendix:experiments}

In this section we specify the hyper parameters used in our experiments and our method for hyper parameter search for both the baselines and double networked averaging agents.

\textbf{Computational Infrastructure}: All experiments use either CPU model Intel i9-9900X (20) @ 4.5GHz with 64GB of RAM running Pop!\_OS 18.04 LTS x86\_64. Or, CPU AMD EPYC 9224 (96) 2.5GHz 72GB of RAM running  Debian (bookworm) x84\_64.

\textbf{Evaluation protocol}: In our experiments we use the same performance metrics as ~\cite{papoudakis_2021}, which consists of periodically stop training for an evaluation checkpoint. For the {\em on-policy } algorithms we use 20 million timesteps except Tag that requires 40 million timesteps to train. For the {\em off-policy} algorithms we use 5 million timesteps for all tasks. There is a total of forty one evaluation checkpoints each of which consisting of running 100 episodes for each random seed and recording the average return obtained across seeds.

\textbf{Baselines}: This work extends the code base from reference~\cite{papoudakis_2021}, particularly the implementations for two baseline algorithms, MAA2C and VDN, from its repository\footnote{\texttt{\url{https://github.com/uoe-agents/epymarl}}}. The implementations from the independent learners INDA2C and INDQL are adaptations from IA2C and IQL respectively. The difference is that the former pair learn from factored rewards while the latter learns from joint rewards. The distributed-V~\cite{zhang_2018} more closely relates to the IA2C implementation from each we adopt the hyperparameters reported by~\citet{papoudakis_2021}. The permutation invariant critic PIC~\cite{liu_2020} more closely relates to MADDPG~\cite{lowe_2017} from which we adopt the hyperparameters reported by~\citet{papoudakis_2021}. 

\textbf{Hyperparameters}: To ensure fairness we did not perform hyperparameter optimization for double networked averaging agents except on three particular hyperparameters $K$ the number of consensus steps, $I$ interval between parameter consensus steps, and $C$ the fixed number of edges on each active edge set $E_k$. We report on the exact hyperparameter values in Tables~\ref{tab:onpolicy-hyperparameter-baselines} and~\ref{tab:hyperparameter}. Moreover the hyperparameters can be divided into three categories: Common, belonging to a family of algorithms,\eg, actor-critic, or to a single implementation of algorithm, \eg, PIC. The hyperparameters' descriptions are:
    \begin{itemize}
        \item  {\textbf{hidden dimension}} The number of neurons in the hidden layer of the neural networks. 
        \item  {\textbf{learning rate}} Regulates the step size of the gradient updates.
        \item  {\textbf{reward standardization}} Performs reward normalization.
        \item  {\textbf{reward type}} Either factored reward, or joint reward.
        \item  {\textbf{network type}} Feed forward and fully connected (FC) or Gated Recurrent Unit (GRU).
        \item {\textbf{entropy coefficient}} (actor-critic family)  The entropy coefficient controls the amount of entropy regularization on  actor's loss function. Entropy regularization facilitates the generation of stochastic policies preventing the action distribution,  at any given State, to over commit to a single optimal action.
        \item  {\textbf{target update}} In 0.01 (soft) mode the {\em target network} is updated with parameters from {\em behavior network} every training step, following a exponentially weighted moving average, with innovation rate 0.01. In 200 (hard) mode the target network is updated with a full copy from the behavior  network at every 200 training steps.
        \item  {\textbf{n-step}} The number of episode steps used to estimate the {\em  discounted return}.
        \item  {\textbf{evaluation epsilon}} ($\epsilon$-greedy family) Epsilon is a hyperparameter controlling the sampling of sub-optimal actions from $Q$-value based policies. The {\em epsilon greedy} criteria provides a way for the agent to experiment with non-greedy actions  actions to find more profitable states. Evaluation epsilon regulates the rate of non-greedy actions taken during evaluation checkpoints. 
        \item  {\textbf{epsilon anneal}} ($\epsilon$-greedy family) The number of episode steps to reach the minimum epsilon for $Q$-value based policies.
        \item  {\textbf{regularization}} (PIC algorithm) The parameter regulating policy entropy.
        \item  {\textbf{pool type}} (PIC algorithm)The convolution parameter which determines how to aggregate the features from the $Q$-critic. 
        \item  {\textbf{K}} (networked family) The number of communication rounds within a parameter communicating  training step. For DNA family every training step is a team-$Q$ consensus step -- not necessarily a parameter communicating training step.
        \item  {\textbf{I}} (networked family)The interval between two sequential parameter communicating training step. 
        \item  {\textbf{C}} (networked family) The number of edges in the communication channel.
    \end{itemize}

\begin{table*}
    \caption{Prescribed hyperparameters  for {\em on-policy} baseline evaluations per algorithm and environment. Reference~\protect\cite{papoudakis_2021} tables INDA2C (Table 15), DVA2C (Table 15) and MAA2C (Table 22).}\label{tab:onpolicy-hyperparameter-baselines}
    \begin{center}
        \begin{tabular}{lrrrrrrr}
            \toprule
                                    & \multicolumn{3}{c}{\textbf{LBF}} & &\multicolumn{3}{c}{\textbf{MPE}}\\\cline{2-4}\cline{6-8}
               \textbf{Hyperparameter} &\textbf{INDA2C} & \textbf{DVA2C} & \textbf{MAA2C} &  & \textbf{INDA2C} & \textbf{DVA2C} & \textbf{MAA2C} \\
             \midrule
             hidden dimension & 64 & 64 & 128 & & 128 & 128 & 128\\
             learning rate & 0.0005 & 0.0005 & 0.0005 & & 0.0005 & 0.0005 & 0.0005\\
             reward standardization & True & True &  True & & True & True & True \\
             reward type & Individual & Individual & Joint & & Individual & Individual & Joint\\
             network type & GRU & GRU & GRU & & FC & FC & GRU \\
             n-step & 5 & 5 & 10 & & 10 & 10 & 5  \\
             target update &  0.01 (soft)& 0.01 (soft) & 0.01 (soft) & & 0.01 (soft) & 0.01 (soft) & 0.01 (soft)\\\hline
             entropy coefficient & 0.01 & 0.01 & 0.01 & & 0.01 & 0.01 & 0.01 \\\hline
             K &    - & 5 & - &     &   - & 5 & - \\
             I &  - & 10 & - &      & - & 10 & - \\
             C &     - & 1 & -  &   &    - & 1 & -  \\
            \bottomrule
        \end{tabular}
    \end{center}
\end{table*}

\begin{table*}
    \begin{center}
    \caption{Prescribed hyperparameters  for {\em off-policy} baseline evaluations per algorithm and environment. Reference~\protect\cite{papoudakis_2021} tables INDQL (Table 13), PIC (Table 19) and VDN (Table 26).}\label{tab:offpolicy-hyperparameter-baselines}
        \begin{tabular}{lrrrrrrr}
            \toprule
                                    & \multicolumn{3}{c}{\textbf{LBF}} & &\multicolumn{3}{c}{\textbf{MPE}}\\\cline{2-4}\cline{6-8}
               \textbf{Hyperparameter} & INDQL & PIC  & VDN &  &  INDQL & PIC & VDN  \\
             \midrule
             hidden dimension & 64 & 64 & 128 & & 128 & 128 & 128  \\
             learning rate & 0.0003 & 0.0003 & 0.0003 & & 0.0005 & 0.0005 & 0.0005 \\
             reward standardization & True & True &  True & & True & True & True \\
             reward type    & Individual & Joint & Joint & & Individual & Joint & Joint \\ 
             network type & GRU & FC & GRU & & FC & GRU & FC \\
             n-step & 1 & - & 1 & & 1 & - & 1 \\
             target update  & 200  & 200  & 0.01  & & 0.01  & 200  & 200 \\\hline
             evaluation epsilon & 0.05 & - & 0.00 & & 0.0 & - & 0.0\\
             epsilon anneal time &  125000 &  - & 500000 & & 200000 & - & 50000 \\\hline
             regularization & - & 0.001 & - & & - & 0.001 & - \\
             pool type & - & max & - & & - & max & -\\
            \bottomrule
        \end{tabular}
    \end{center}
\end{table*}

Tables~\ref{tab:onpolicy-hyperparameter-baselines}  and ~\ref{tab:offpolicy-hyperparameter-baselines} report hyperparameter for the baseline algorithms.

\textbf{Hyperparameter Search}: For double networked averaging actor critic with advantage (DNAA2C) we use the hyperparameters reported by~\citep{papoudakis_2021} Table 15. For double networked averaging $Q$-learning (DNAQL), we use the hyperparameters reported by~\citep{papoudakis_2021} Table 12. Holding these parameters fixed, we tested the networked agents specific parameters. We tested the combinations: (i) Number of consensus rounds ($K$): 1/5/10, (ii) training interval ($I$) between two rounds of consensus on the parameters: 1/5/10, (iii) The number of edges on the communication graph ($C$): 1/2. Each hyperparameter combination is evaluated for three different and for 20 million time steps (on-policy) and 40 milion time steps (off-policy). For each combination of hyperparameters, we select two evaluation checkpoints neighborhood around the evaluation checkpoint that received the maximum average episodic return over the three seeds. The hyperparameter combination, represented by its 15 points sample, with maximum average return is selected. Table~\ref{tab:hyperparameter} reports on the resulting hyperparameter combination from the hyperparameter grid-search process.
 
\begin{table}
   \caption{The selected hyperparameters of DNAA2C test evaluations are based from reference~\protect\cite{papoudakis_2021} Table 15. And the selected hyperparameters of DNAQL test evaluation are based from reference~\protect\cite{papoudakis_2021} Table 12.}\label{tab:hyperparameter}
   \begin{center}
        \begin{tabular}{lrrrrr}
            \toprule
            
                                    & \multicolumn{2}{c}{\textbf{LBF}} & &\multicolumn{2}{c}{\textbf{MPE}}\\\cline{2-3}\cline{5-6}
               \textbf{Hyperparameter} & \textbf{DNAA2C} &\textbf{DNAQL} & & \textbf{DNAA2C} & \textbf{DNAQL}\\
             \midrule
             hidden dimension  & 64  & 128 & & 128 & 128\\
             learning rate  & 0.0005  & 0.003 &  &  0.0005 & 0.0003 \\
             reward standardization  & True  & True & &  True & True\\
             reward type & Independent & Independent & & Independent & Independent\\
             network type   & GRU  & GRU & & GRU & FC\\
             n-step    &  5 &  5 &  & 10 & 5\\
             target update  &  0.01(soft)  &  200 (hard) &  &   0.01 (soft) &   0.01 (soft)\\\hline
             entropy coefficient & 0.01  &  - &  & 0.01  & -   \\\hline
             evaluation epsilon &  - &   0.0 &   &  -  & 0.0\\
             epsilon anneal time &  - & 500000 &  & - & 500000\\\hline
             K &  5  & 1 &  &   5  & 1\\
             I & 10 & 10 &  & 10  & 1\\
             C & 1   &  1 &  &  1  & 1\\
            \bottomrule
        \end{tabular}
   \end{center}
\end{table}

\section{Extended Results}\label{appendix:results}
    
\begin{figure}
    \captionsetup{skip=20pt}
     \begin{center}
          \begin{subfigure}{0.24\linewidth}
                  \includegraphics[width=\linewidth]{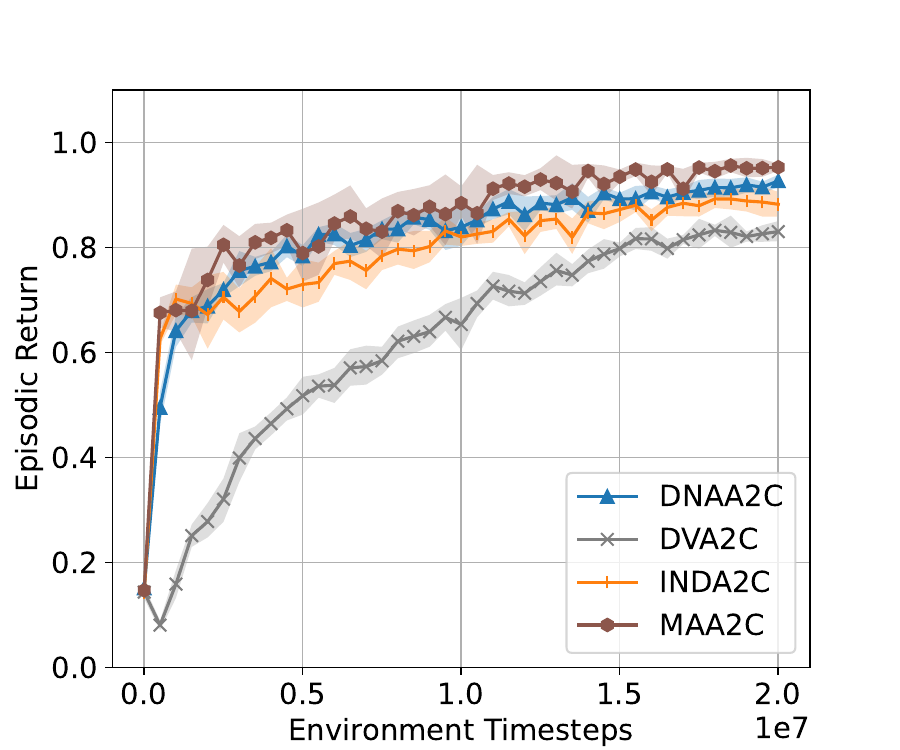}
                  \caption{LBF: Easy instance}
          \end{subfigure}
          \begin{subfigure}{0.24\linewidth}
                  \includegraphics[width=\linewidth]{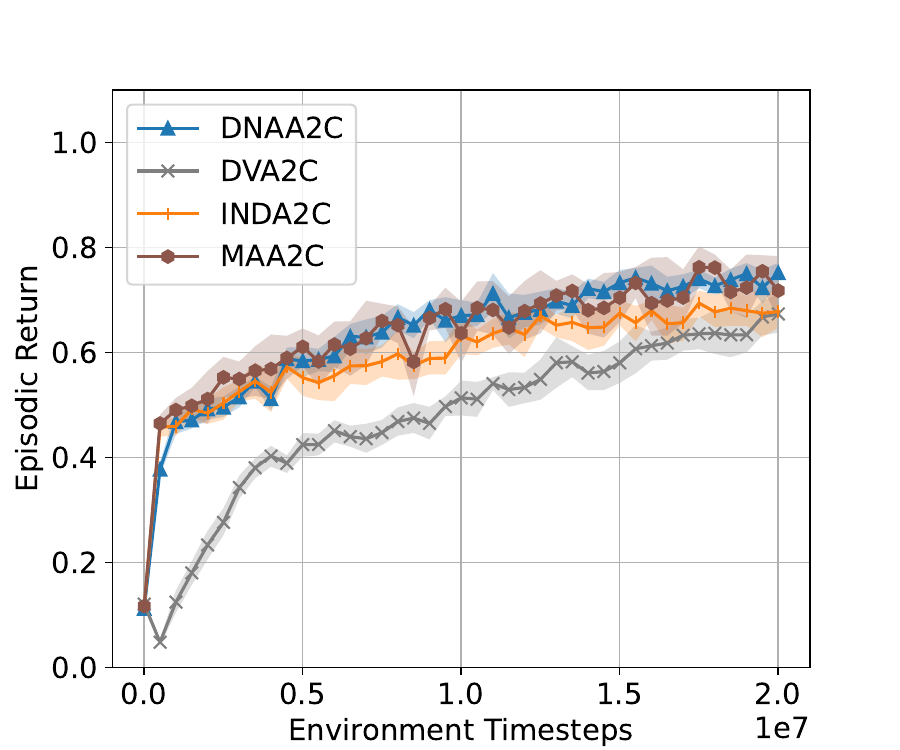}
                  \caption{LBF: Medium instance}
          \end{subfigure}
          \begin{subfigure}[b]{0.24\textwidth}
                \includegraphics[width=\linewidth]{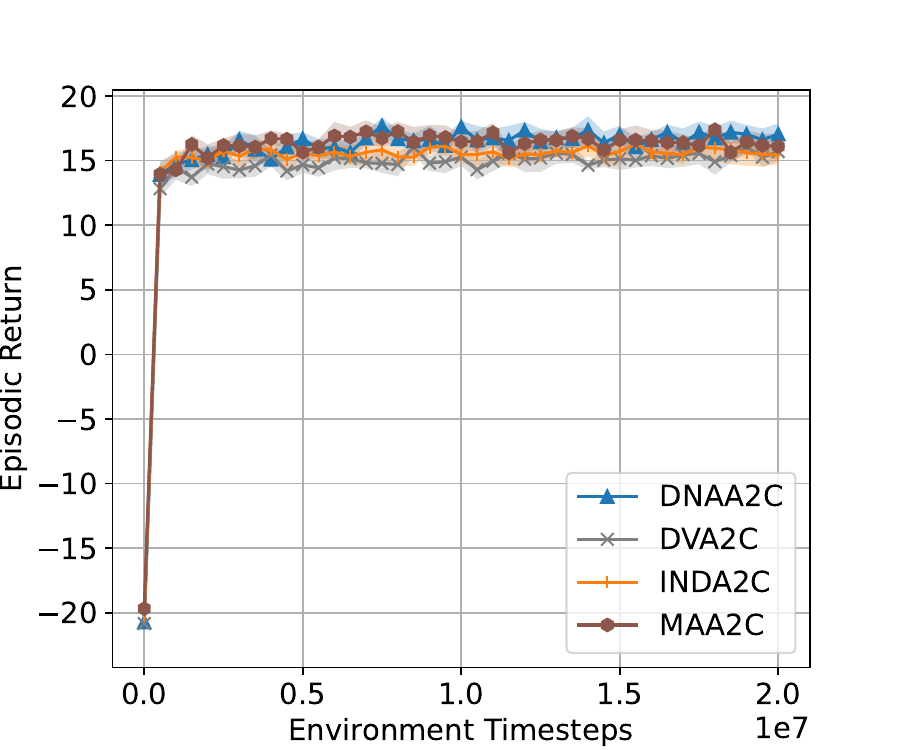}
                \caption{MPE: Adversary}
          \end{subfigure}
          \begin{subfigure}[b]{0.24\textwidth}
                \includegraphics[width=\linewidth]{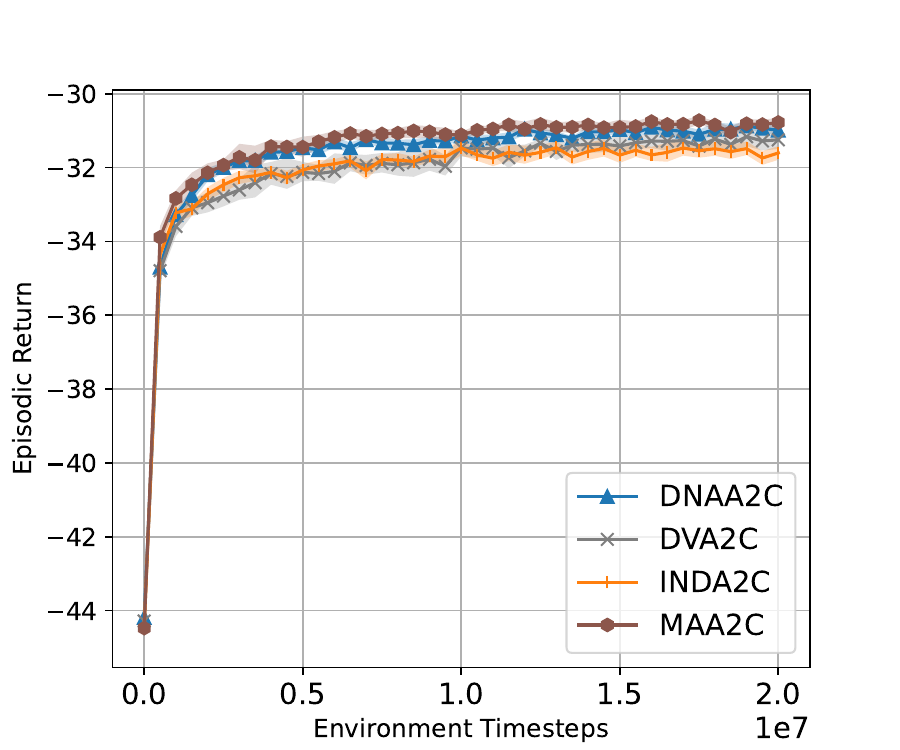}
                \caption{MPE: Spread}
          \end{subfigure}
      \end{center}
  \caption{Train rollouts for the {\em on-policy} algorithms. The bullets represent the average of evaluation checkpoints for ten random seeds. The shaded area represents a 95\% bootstrap confidence interval around the average. In blue DNAA2C (ours), in orange INDA2C, the independent agents system. In gray, DVA2C, a distributed-$V$ algorithm~\protect \cite{zhang_2018}. In chestnut MAA2C, the central-$V$, acts as an upper bound for performance. For the four instances, DNAA2C provides the best approximation for MAA2C.}\label{fig:onpolicy} 
    \Description{Train rollouts for the additional on-policy setting tasks: Easy and Medium tasks (LBF environment). And Adversary and Spread tasks (MPE environment).}
\end{figure}

\begin{figure}
    \begin{center}
        \begin{subfigure}{0.24\linewidth}
            \includegraphics[width=\linewidth]{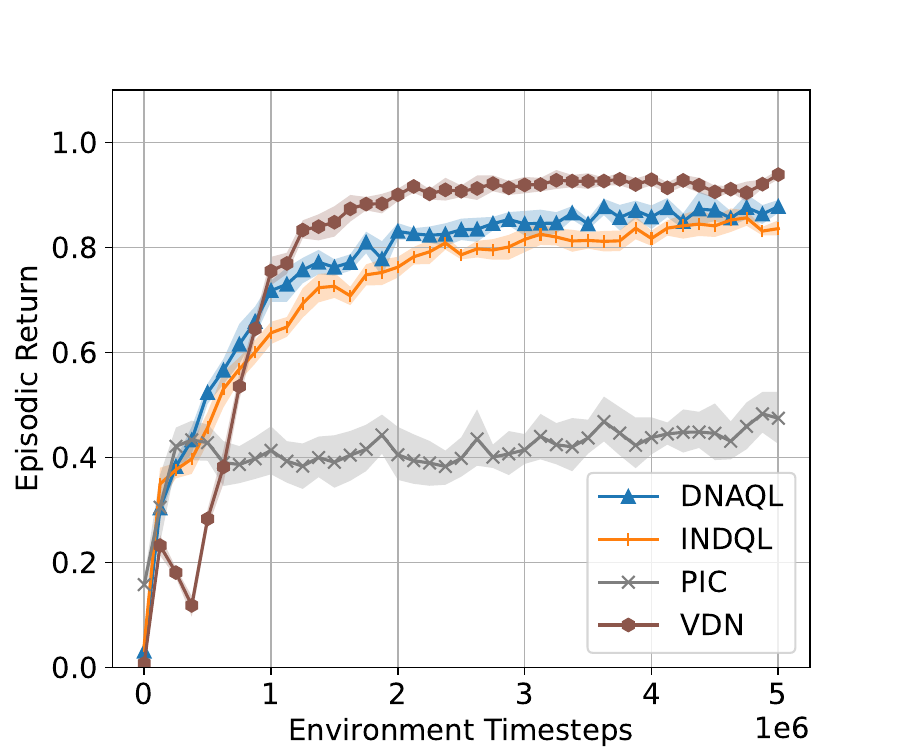}
            \caption{LBF: Easy}
        \end{subfigure}
        \begin{subfigure}{0.24\linewidth}
             \includegraphics[width=\textwidth]{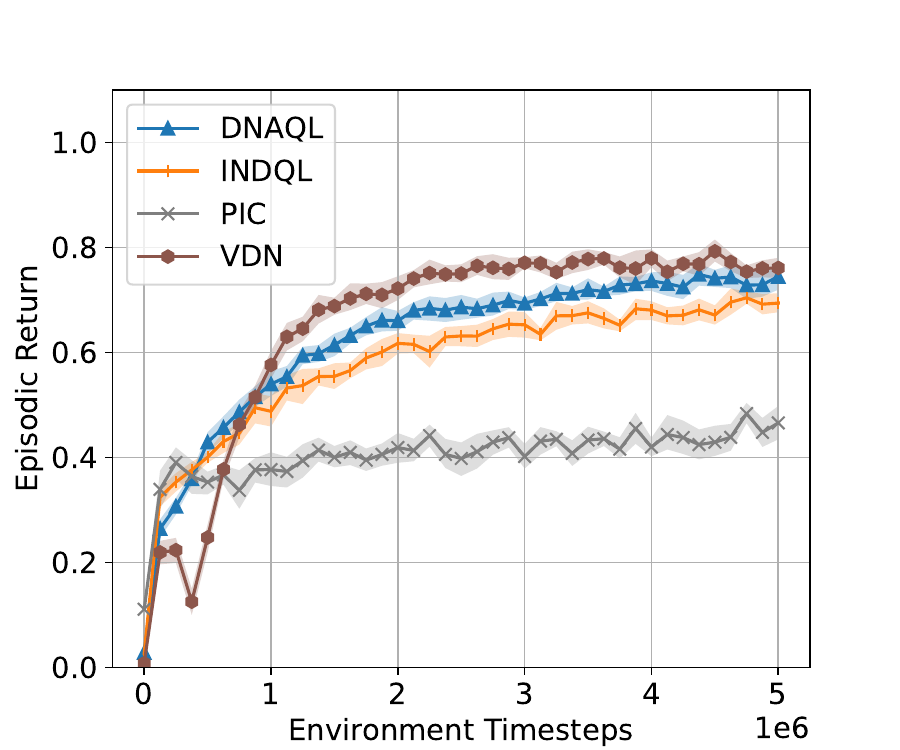}
             \caption{LBF: Medium}
        \end{subfigure}
        \begin{subfigure}[b]{0.24\textwidth}
            \includegraphics[width=\textwidth]{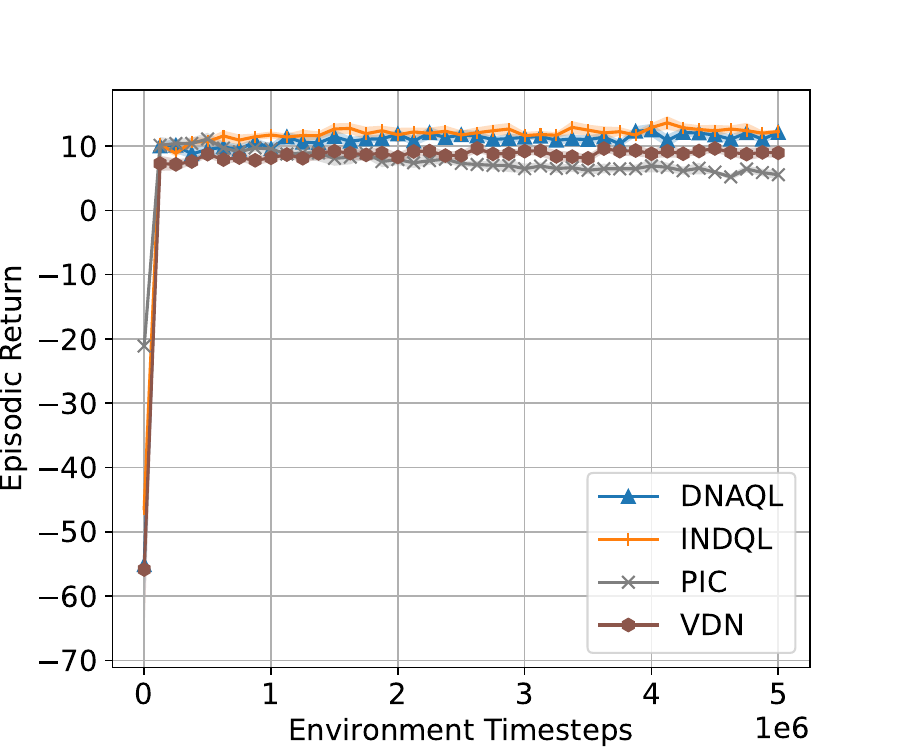}
            \caption{MPE: Adversary}
        \end{subfigure}
        \begin{subfigure}[b]{0.24\textwidth}
            \includegraphics[width=\textwidth]{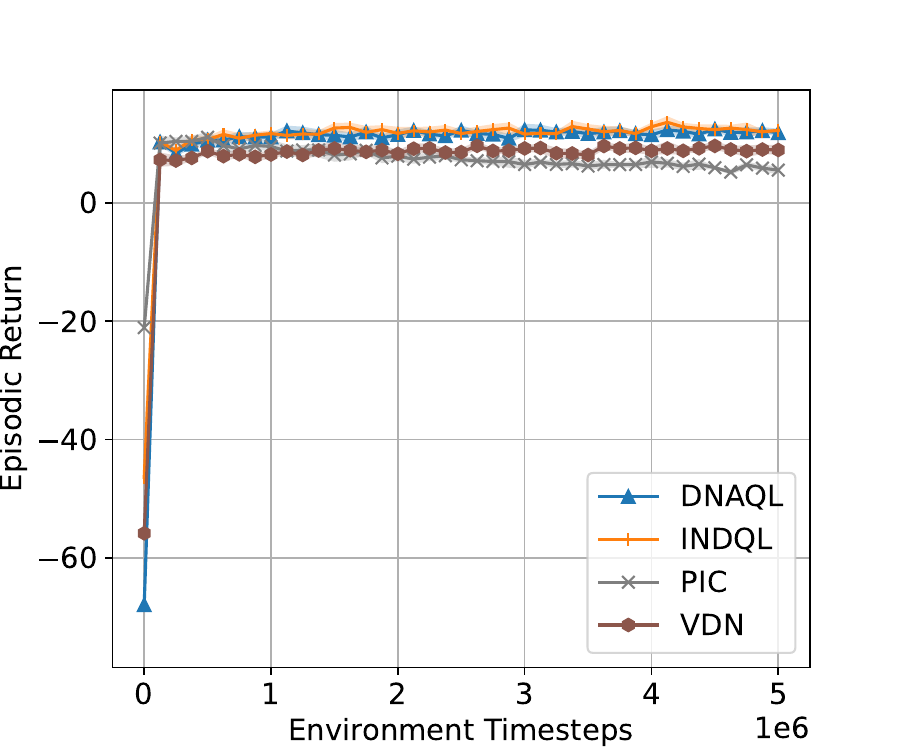}
            \caption{MPE: Spread}
        \end{subfigure}
    \end{center}
  \caption{Train rollouts for the {\em off-policy} algorithms. The bullets represent the average of evaluation checkpoints for ten random seeds. The shaded area represents a 95\% bootstrap confidence interval around the average. In blue DNAQL (ours), in orange INDQL, the independent agents system. In gray, PIC a central-$Q$ algorithm that approximates~\protect \cite{chen_2022}. In chestnut VDN, that factorized representations for a central $Q$. For the LBF scenarios DNAQL outperforms other decentralized approaches. For MPE scenarios, (d) and (e), it outperforms VDN. }\label{fig:offpolicy} 
    \Description{Train rollouts for the additional on-policy setting tasks: Easy and Medium tasks (LBF environment). And Adversary and Spread tasks (MPE environment).}
\end{figure}

\textbf{Performance:} The performance plots for the on-policy algorithms are shown in Figure~\ref{fig:onpolicy}. Across all six tasks, DNAA2C outperforms other decentralized approaches. The performance plots for the off-policy algorithms are shown in Figure~\ref{fig:offpolicy}. In the LBF environment, DNAQL outperforms other decentralized approaches. However, in the MPE tasks Adversary and Spread, INDQL outperforms VDN being the best performing algorithm, followed by DNAQL. Conversely, in the Tag task, VDN outperforms other approaches.

\textbf{Ablations:} To compute the ablation plots, for each ablation group, we identify the maximum average return checkpoint across five random seeds. We extend the selection for a size-2 neighborhood around maximum average return checkpoint. From the resulting the sample of 25 points, we build an empirical distribution by re-sampling those points ten thousand times, and then build a 95\% bootstrap confidence interval. 

\begin{figure}
    \begin{subfigure}{0.24\linewidth}
        \includegraphics[width=\linewidth]{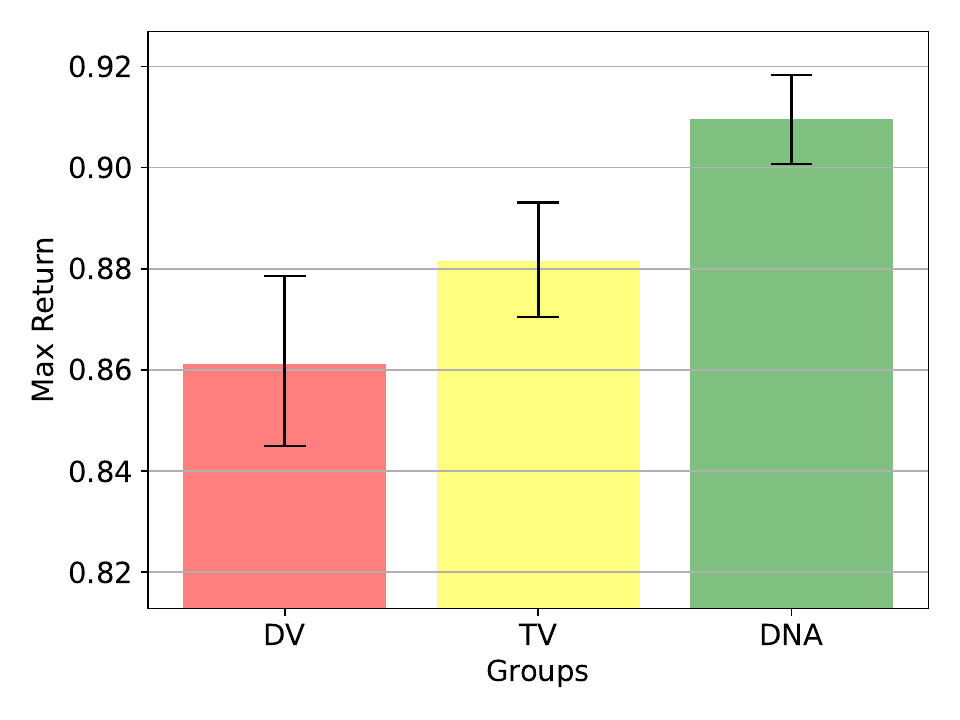}
        \caption{LBF: Easy}
    \end{subfigure}
    \begin{subfigure}{0.24\linewidth}
        \includegraphics[width=\linewidth]{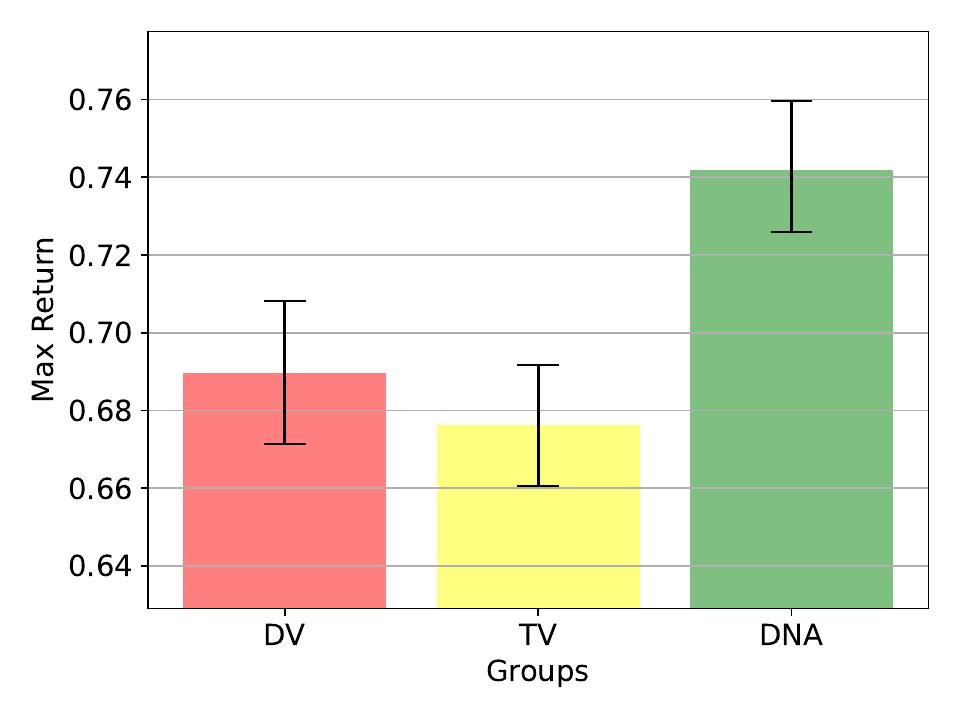}
        \caption{LBF: Medium}
    \end{subfigure}
    \begin{subfigure}{0.24\linewidth}
        \includegraphics[width=\linewidth]{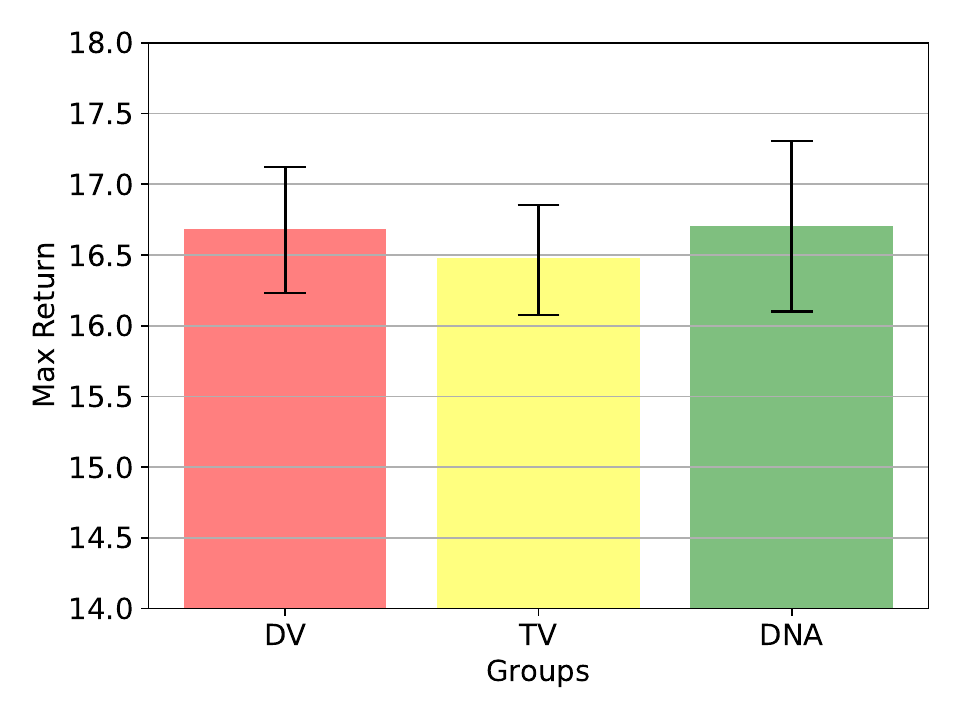}
        \caption{MPE: Adversary}
    \end{subfigure}
    \begin{subfigure}{0.24\linewidth}
        \includegraphics[width=\linewidth]{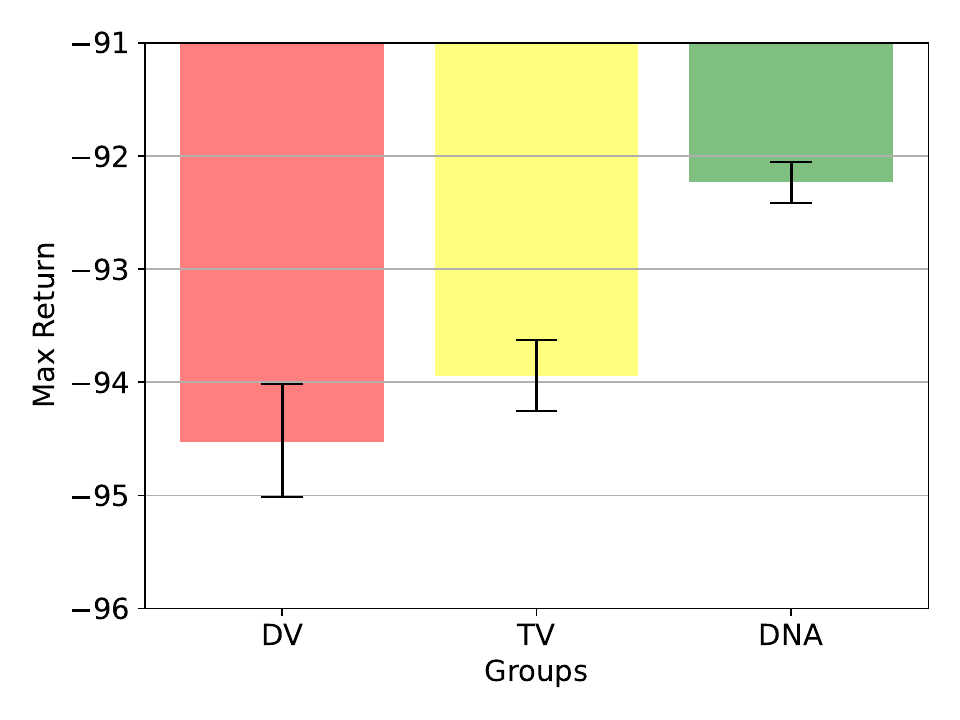}
        \caption{MPE: Spread}
    \end{subfigure}
    \caption{DNAA2C ablations: DV (distributed-V) group learns from individual rewards, performs consensus rounds on the critic. TV (team-V) group performs team-$V$ consensus  (our contribution) and critic consensus. DNA group has team-$V$ consensus, both actor and critic consensus. Team-$V$ consensus improves results in tasks LBF: easy and MPE: Spread. DNA improves three from four tasks.}\label{fig:onpolicy_ablation}
\end{figure}

There are three modules to the DNAA2C algorithm: team-$V$ consensus aimed at cooperation, critic consensus aimed at an agreement on a single critic for the joint policy and the policy consensus which emulates parameter sharing. Figure~\ref{fig:onpolicy_ablation} plots the three relevant groups;the distributed-$V$ (DV) group that has only the critic consensus turned on; the team-$V$ (TV) group that has target consensus and critic consensus turned on; and DNA group, which has target consensus and parameter consensus turned on the actor and critic. We note that TV group improves the DV group in two tasks and that DNA improves three out of four tasks.

\begin{figure}
    \hfill
    \centering
    \begin{subfigure}{0.24\linewidth}
        \includegraphics[width=\linewidth]{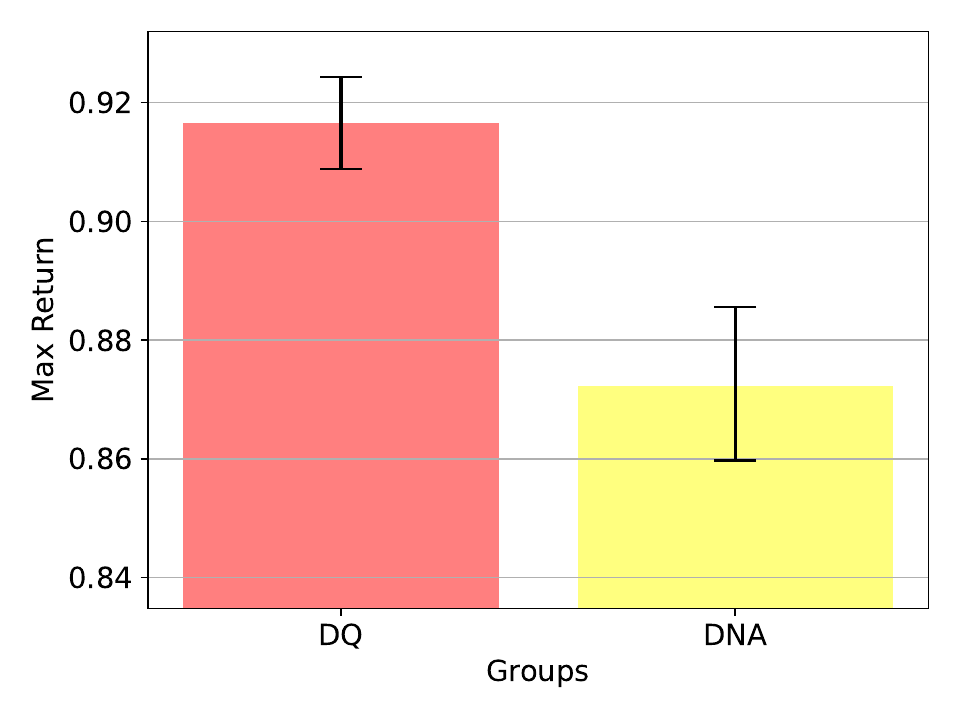}
        \caption{LBF: Easy}
    \end{subfigure}
    \begin{subfigure}{0.24\linewidth}
        \includegraphics[width=\linewidth]{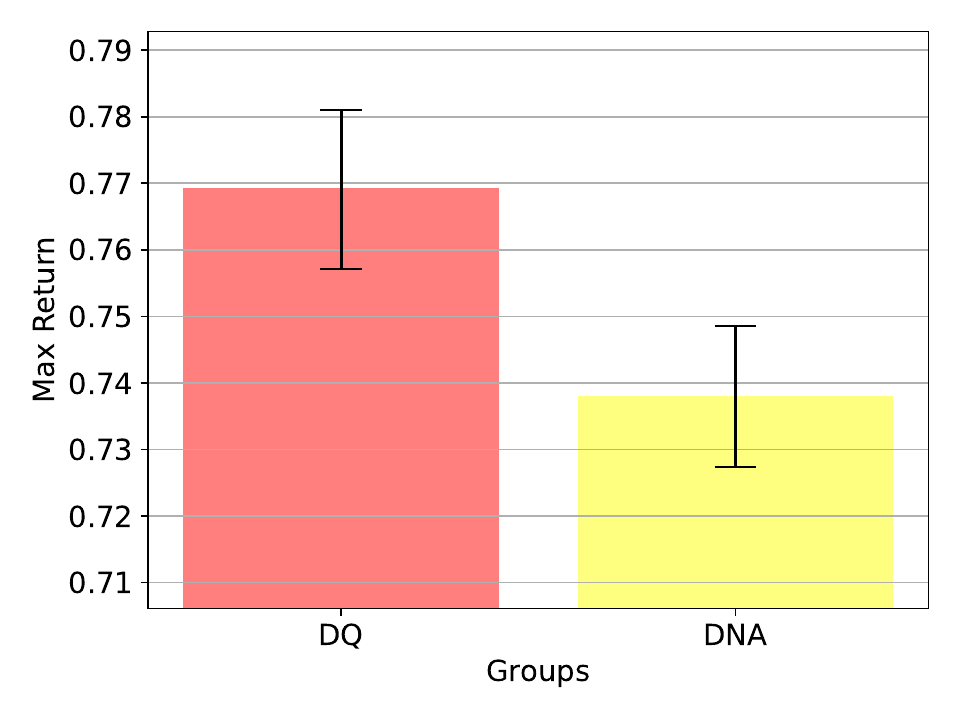}
        \caption{LBF: Medium}
    \end{subfigure}
    \begin{subfigure}{0.24\linewidth}
        \includegraphics[width=\linewidth]{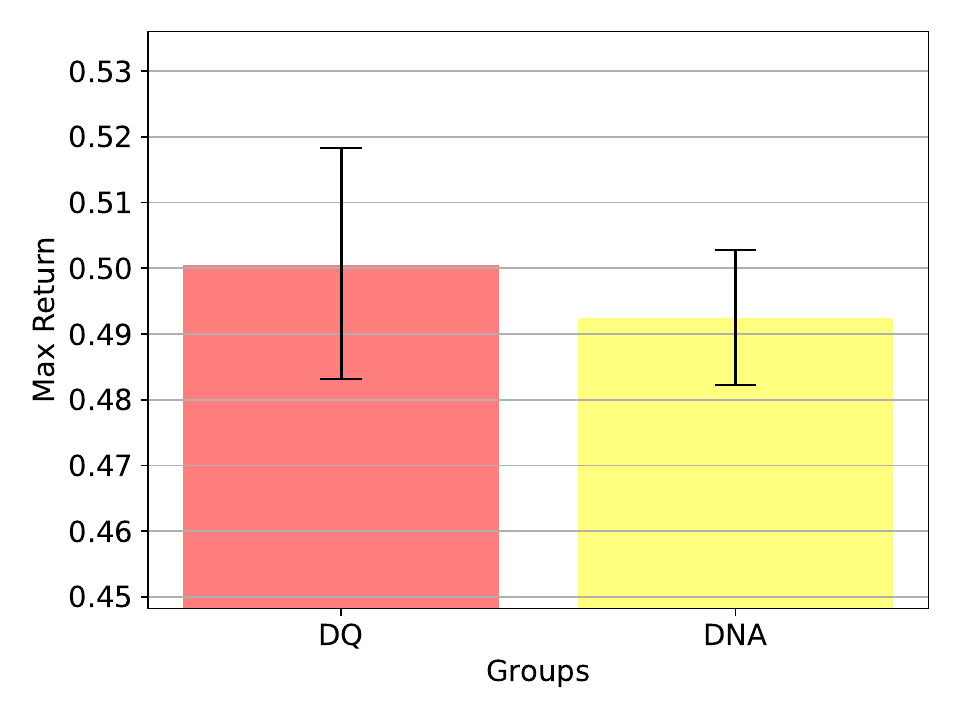}
        \caption{LBF: Hard}
    \end{subfigure}
    \begin{subfigure}{0.24\linewidth}
        \includegraphics[width=\linewidth]{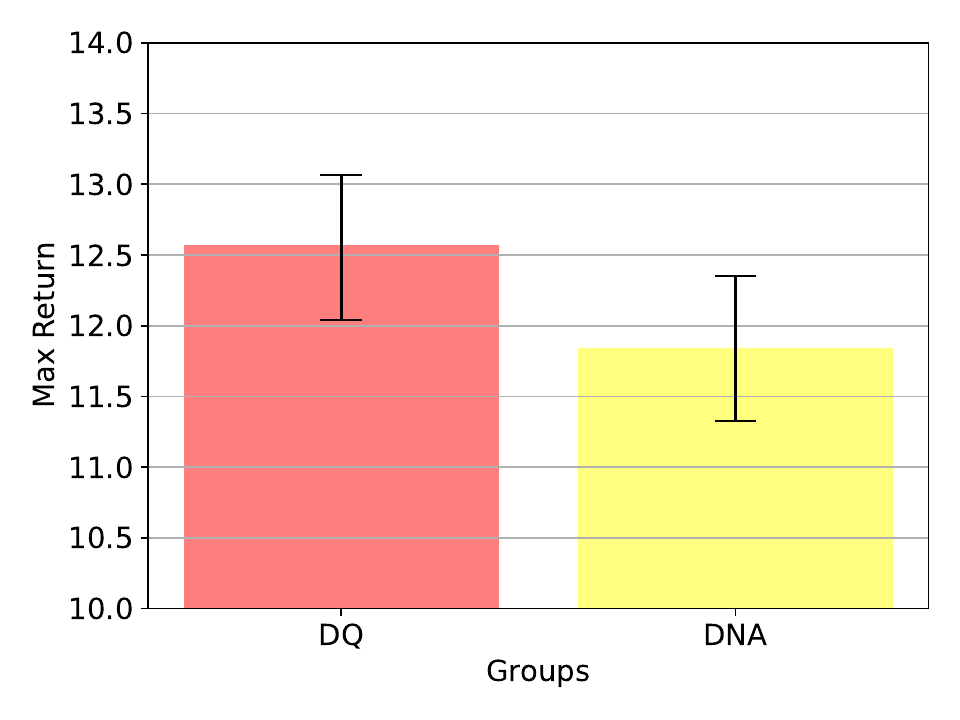}
        \caption{MPE: Adversary}
    \end{subfigure}
    \begin{subfigure}{0.24\linewidth}
        \includegraphics[width=\linewidth]{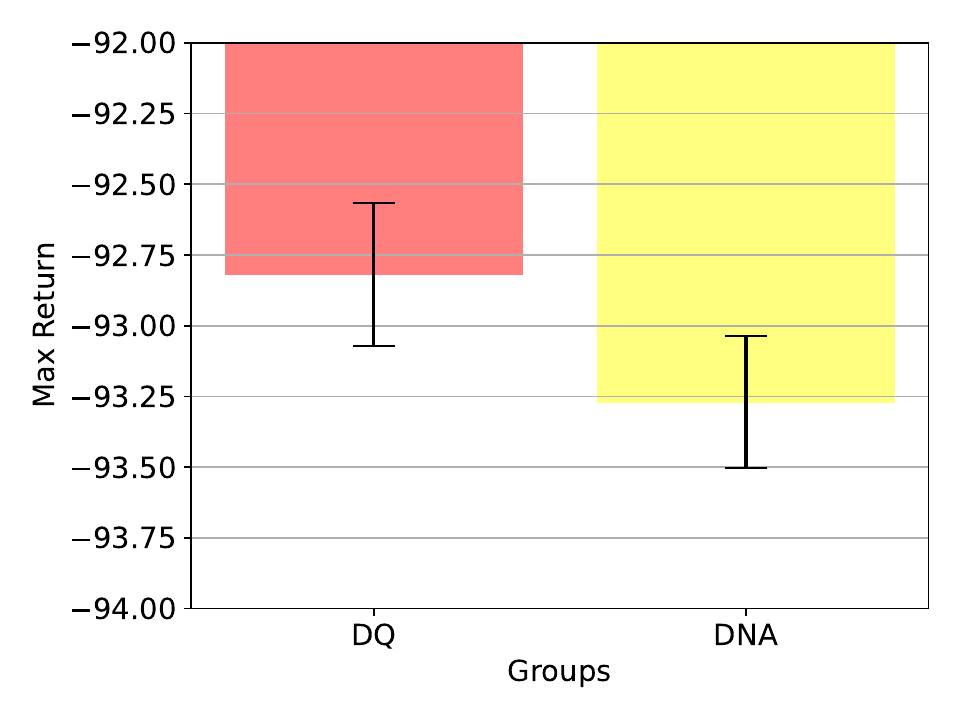}
        \caption{MPE: Spread}
    \end{subfigure}
    \begin{subfigure}{0.24\linewidth}
        \includegraphics[width=\linewidth]{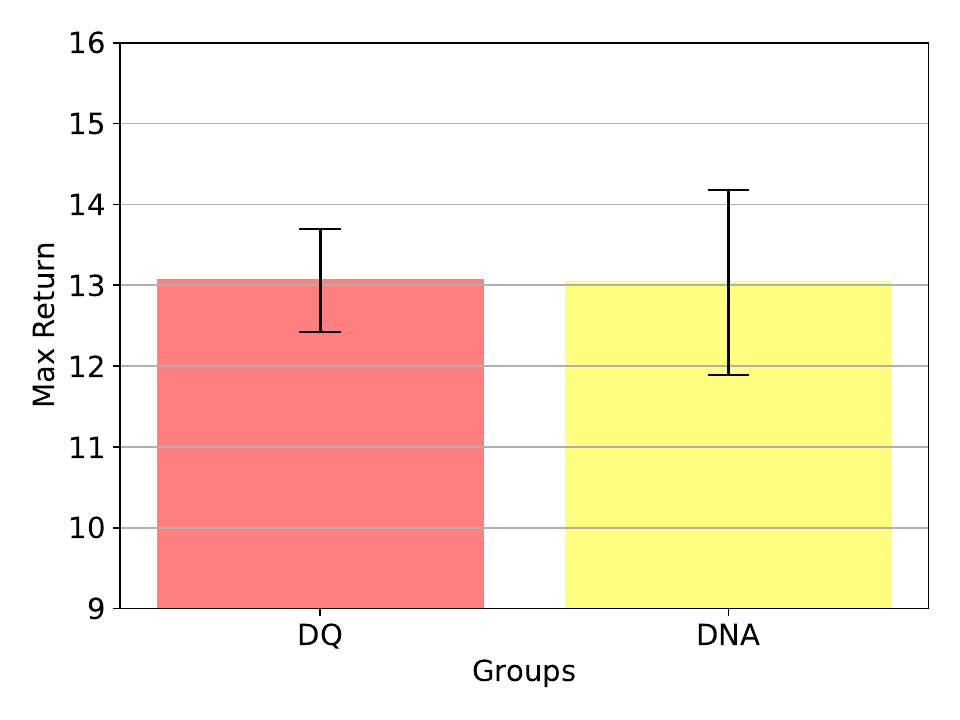}
        \caption{MPE: Tag}
    \end{subfigure}
  \caption{DNAQL ablations: DQ (distributed-$Q$) performs consensus on parameters only. DNA group has both team-$Q$ consensus and parameter consensus. For six tasks the component that improves performance is consensus on the parameters and not team-$Q$ consensus.}\label{fig:offpolicy_ablation} 
\end{figure}

There are two modules to the DNAQL algorithm: team-$Q$ consensus aimed at cooperation and $Q$-critic consensus that plays the dual role of evaluating the $Q$-function and emulates parameter sharing in the distributed setting. Figure~\ref{fig:offpolicy_ablation} plots the two relevant groups, namely, the distributed-$Q$ (DQ) group that has $Q$ consensus turned on, and DNA group that has team-$Q$ consensus and $Q$ consensus. Differently from the on-policy setting, team-$Q$ degrades the performance in the LBF environment, as evidenced by the difference in performance from the DQ and DNA group. We hypothesize that the best $Q$ policy has the agents acting independently: the agents learn how to walk to the nearest fruit and attempt to load it, oblivious to the presence of other agents. Since some are successful, they have the chance to navigate to another fruit and attempt a load action, perhaps with the help of a another agent. For the tasks in MPE environment, the differences in results are small, but the main factor contributing to the result is the DQ.

\section{Extended Related Work}\label{appendix:related_work}

\textbf{Partial observability}: Three approaches to mitigate the negative effects of partial observability in MARL are: The use of memory-based policies, the estimation of multi-agent belief, and the communication of observations. For example,~\cite{omidshafiei_2017} propose ILs that use RNNs to improve agents' observability by conditioning the policies on the history of individual observations.~\cite{peti_2023} propose a DTDE MARL system under partial observability where agents perform consensus iterations over the multi-agent belief, a joint distribution over states and joint actions. But since the complexity grows exponentially with the number of states their system is not scalable. ~\cite{wang_2020} propose a CTDE MARL system that develops a memory-based actor-critic where agents learn when to communicates their own position. The DNA-MARL approach is model-free and it does not require the estimation of a multi-agent belief. Finally, this approach differs from others in that it does not require agents to control when to communicate, or share their own observations.

\textbf{Learning cooperation} (LC):~\citet{hernandez-leal_2019} point this research thread as one of the  main paradignms in MARL literature. LC can be sub divided into three topics: (i) Learning how to cooperate in social dilemmas, (ii) Improving deep MARL under the fully cooperative setting and (iii) And MARL in mixed settings. 

First, learning how to cooperate in social dilemmas relates to emergent cooperative behavior in the presence of conflicting interests.  This branch doesn't directly compare to our work, for instance,~\cite{stimpson_2003, decote_2006} are equivalent to one-state sequential games and are closely related to bandit algorithms~\cite{lattmore_2020}.  In DNA-MARL, agents receive an individual reward, but attempt to generate a team value evaluation by sharing the their value function values.

The second type of work slightly modify the objective function to improve learning in cooperative settings. For instance {\em hysterenic $Q$-learning}~\cite{omidshafiei_2017} changes the well known $Q$-learning update, to discount exploratory actions from teammates. Similarly, {lenient learning}~\cite{potter_1994, palmer_2018} also apply optimistic updates to $Q$-learning preventing a pathology in learning known as {\em relative over generalization}.

The third type of work, MARL in mixed settings are algorithms that learn from a individual reward, \eg,~\cite{lowe_2017, liu_2020}. DNA-MARL is suitable to settings where nature draws individual rewards but we would like to design a cooperative system do solve a distributed problem. For instance, in adaptive traffic signal control task, agents only observe local traffic (individual rewards), but must cooperate to control network traffic. 

\textbf{Partially observable Markov game}: Markov game (MG)~\cite{littman_1994} is the framework underpinning multi-agent reinforcement learning. MG is the generalization to the Markov decision processes~\cite{sutton_2018} whereby multiple agents interact sequentially with a shared environment to maximize their individual sum of rewards. Our work is an extension from networked multi-agent Markov decision process~\cite{zhang_2018} to the partial observability setting.~\citet{zhang_2018} develop two cooperative algorithms both of which have full observability with respect to the state space and the joint action, but collect individual rewards. The first solution involves estimating a local advantage function $A^i_\theta(s,a)$. The local advantage function requires that each agent approximates a global $Q$-function that represent both the state and joint action space, thereby exacerbating the challenge posed by the large state space.~\citet{cassano_2019} report that this solution, which depends on \em{Nash equilibria}~\cite{littman_1994}, may become stuck in a \em{sub-optimal Nash equilibrium}~\cite{gronauer_2022}. The second solution uses a neural network to approximate the the team reward $\bar{R}(s, a)$ locally. In general, both approaches are suitable when $s$ and $a$ are known, but may fail under partial observability. DNA-MARL uses consensus with respect to the team value~\eqref{eqn:V_consensus}, which provides superior performance in on-policy settings according to our experimental section.

\citet{chen_2022} define a homogeneous Markov game where the state space is jointly fully observable. Jointly fully observability is an amenable form of partial observability~\cite{oliehoek_2016}. To compensate for partial observability their system trains a bandit algorithm~\cite{lattmore_2020} that learns when and to whom communicate, allowing agents to solicit observation and actions from their neighbors. DNA-MARL, however, operates under stricter constrains: agents lack the ability to choose when or to whom communicate, and they do not share local information. This design choice makes DNA-MARL adaptable to domains that require privacy. Despite using extra information, empirical results show that DNAQL outperforms PIC (the representative baseline). 

More recently, there is a class of model-based works that address partial observability by explicitly performing consensus on a {\em multi-agent belief}. Multi-agent belief is a joint probability distribution induced by an initial state distribution, and a history of observations and joint actions~\cite{oliehoek_2016}.~\citet{peti_2023} use average consensus to compel agents to agree on a multi-agent belief. Their system requires {\em exact} representation of the multi-agent belief, its computational complexity grows exponentially on the states and the number of agents.~\citet{kayaalp_2023} use function approximation for multi-agent belief computation. However, their approach focuses on the {\em policy evaluation problem} whereas DNA-MARL develop the broader class of {\em policy evaluation} and {\em policy improvement}~\cite{sutton_2018}.

\end{document}